\documentclass[11pt]{article}

\usepackage[final]{acl}
\usepackage{hyperref}
\usepackage{url}
\usepackage[ruled,vlined]{algorithm2e} 
\usepackage{graphicx}
\usepackage{enumitem}
\usepackage{float}
\usepackage{booktabs}
\usepackage{tabularx}
\usepackage[HTML, table]{xcolor}
\definecolor{MyPurple}{HTML}{9933FF}
\usepackage{times}
\usepackage{latexsym}
\usepackage{booktabs}   
\usepackage{tabularx}   
\usepackage[table]{xcolor} 
\usepackage[T1]{fontenc}

\usepackage[utf8]{inputenc}

\usepackage{microtype}

\usepackage{inconsolata}

\usepackage{xcolor}

\usepackage{graphicx}
\usepackage{amsmath}
\usepackage{amssymb}
\usepackage{cleveref}
\newcount\heata     
\newcount\heatb     
\newcount\heatden   
\newcount\heattmp
\newcount\heatpct
\newcount\heatval

\definecolor{orange}{RGB}{255,180,129}
\definecolor{lightblue}{RGB}{200,235,255}

\makeatletter
\def\heatparsecount#1#2{%
  \edef\heatta{#2}%
  \expandafter\heatgetint\heatta.\@nil
  #1=\numexpr\heatint\relax
}
\def\heatgetint#1.#2\@nil{\def\heatint{#1}}
\makeatother

\newcommand{\setheatlimits}[2]{%
  \heatparsecount\heata{#1}%
  \heatparsecount\heatb{#2}%
  \def\heatas{#1}%
  \def\heatbs{#2}%
}

\makeatletter
\def\heatsetval#1{%
  \edef\heatta{#1}%
  \expandafter\heatgetint\heatta.\@nil
  \heatval=\numexpr\heatint\relax
}
\makeatother

\newcolumntype{Y}[2]{>{\setheatlimits{#1}{#2}}r}

\newcommand{\heat}[1]{%
  \begingroup
  \heatsetval{#1}%
  \heatden=\numexpr\heatb-\heata\relax
  \ifnum\heatden=0
    \heatpct=50 
  \else
    \heattmp=\numexpr\heatval-\heata\relax
    \ifnum\heatden>0
      \ifnum\heattmp<0 \heattmp=0 \fi
      \ifnum\heattmp>\heatden \heattmp=\heatden \fi
    \else
      \ifnum\heattmp>0 \heattmp=0 \fi
      \ifnum\heattmp<\heatden \heattmp=\heatden \fi
    \fi
    \heatpct=\numexpr(100*\heattmp)/\heatden\relax 
    \ifnum\heatpct<0 \heatpct=0 \fi
    \ifnum\heatpct>100 \heatpct=100 \fi
  \fi
  \edef\colorspec{lightblue!\the\heatpct!orange}%
  \expandafter\cellcolor\expandafter{\colorspec}{#1}%
  \endgroup
}

\newcommand{\heatlog}[1]{%
  \begingroup
  \heatsetval{#1}%
  \heatden=\numexpr\heatb-\heata\relax
  \ifnum\heatden=0
    \heatpct=50
  \else
    \ifnum\heatden>0
      \ifnum\heatval<\heata \heattmp=\heata \else
      \ifnum\heatval>\heatb  \heattmp=\heatb  \else \heattmp=\heatval \fi \fi
    \else
      \ifnum\heatval>\heata \heattmp=\heata \else
      \ifnum\heatval<\heatb  \heattmp=\heatb  \else \heattmp=\heatval \fi \fi
    \fi
    \ifnum\heata>0 \ifnum\heatb>0 \ifnum\heattmp>0
      \edef\heatvc{\the\heattmp}%
      \edef\percstr{\fpeval{round(100*(ln(\heatvc)-ln(\heatas))/(ln(\heatbs)-ln(\heatas)),0)}}%
      \heatpct=\numexpr\percstr\relax
      \ifnum\heatpct<0 \heatpct=0 \fi
      \ifnum\heatpct>100 \heatpct=100 \fi
    \else\else\else
      \heattmp=\numexpr\heatval-\heata\relax
      \ifnum\heatden>0
        \ifnum\heattmp<0 \heattmp=0 \fi
        \ifnum\heattmp>\heatden \heattmp=\heatden \fi
      \else
        \ifnum\heattmp>0 \heattmp=0 \fi
        \ifnum\heattmp<\heatden \heattmp=\heatden \fi
      \fi
      \heatpct=\numexpr(100*\heattmp)/\heatden\relax
      \ifnum\heatpct<0 \heatpct=0 \fi
      \ifnum\heatpct>100 \heatpct=100 \fi
    \fi\fi\fi
  \fi
  \edef\colorspec{lightblue!\the\heatpct!orange}%
  \expandafter\cellcolor\expandafter{\colorspec}{#1}%
  \endgroup
}

\newcommand{\heatlogp}[1]{%
  \begingroup
  \heatsetval{#1}%
  \heatden=\numexpr\heatb-\heata\relax
  \ifnum\heatden=0
    \heatpct=50
  \else
    \ifnum\heatden>0
      \ifnum\heatval<\heata \heattmp=\heata \else
      \ifnum\heatval>\heatb  \heattmp=\heatb  \else \heattmp=\heatval \fi \fi
    \else
      \ifnum\heatval>\heata \heattmp=\heata \else
      \ifnum\heatval<\heatb  \heattmp=\heatb  \else \heattmp=\heatval \fi \fi
    \fi
    \ifnum\numexpr\heata+1\relax>0 \ifnum\numexpr\heatb+1\relax>0
      \edef\heatvc{\the\heattmp}%
      \edef\percstr{\fpeval{round(100*(ln(1+\heatvc)-ln(1+\heatas))/(ln(1+\heatbs)-ln(1+\heatas)),0)}}%
      \heatpct=\numexpr\percstr\relax
      \ifnum\heatpct<0 \heatpct=0 \fi
      \ifnum\heatpct>100 \heatpct=100 \fi
    \else\else
      \heattmp=\numexpr\heatval-\heata\relax
      \ifnum\heatden>0
        \ifnum\heattmp<0 \heattmp=0 \fi
        \ifnum\heattmp>\heatden \heattmp=\heatden \fi
      \else
        \ifnum\heattmp>0 \heattmp=0 \fi
        \ifnum\heattmp<\heatden \heattmp=\heatden \fi
      \fi
      \heatpct=\numexpr(100*\heattmp)/\heatden\relax
      \ifnum\heatpct<0 \heatpct=0 \fi
      \ifnum\heatpct>100 \heatpct=100 \fi
    \fi\fi
  \fi
  \edef\colorspec{lightblue!\the\heatpct!orange}%
  \expandafter\cellcolor\expandafter{\colorspec}{#1}%
  \endgroup
}

\setheatlimits{0}{100} 

\usepackage[most]{tcolorbox}
\usepackage{cleveref}

\newtcolorbox{quotebox}[1][]{%
    colback=gray!5,
    colframe=gray!80,
    arc=1mm,
    outer arc=1mm,
    boxrule=0.5pt,
    left=3mm, right=3mm,
    top=2mm, bottom=2mm,
    enhanced,
    #1 
}

\title{GeoRC: A Benchmark for Geolocation Reasoning Chains}

\author{
    \textbf{Mohit Talreja, Joshua Diao, Jim Thannikary James, Radu Casapu,} \\
    \textbf{Tejas Santanam, Ethan Mendes, Alan Ritter, Wei Xu, James Hays} \\
    \texttt{\{mtalreja6,jdiao6,jimjames,rcasapu3,tsantanam,emendes3,hays\}@gatech.edu}, \\
    \texttt{\{alan.ritter, wei.xu\}@cc.gatech.edu} \\
    Georgia Institute of Technology, Atlanta, GA, U.S.A \\
    \href{https://talrejamohit03.github.io/GeoRC/}{Project Home},
\href{https://github.com/talrejamohit03/GeoRC}{GitHub}, \href{https://huggingface.co/datasets/mohit-talreja/GeoRC}{HuggingFace}
}

\begin{document}
\maketitle

\begin{figure*}[h!]
    \centering
    \includegraphics[width=\textwidth]{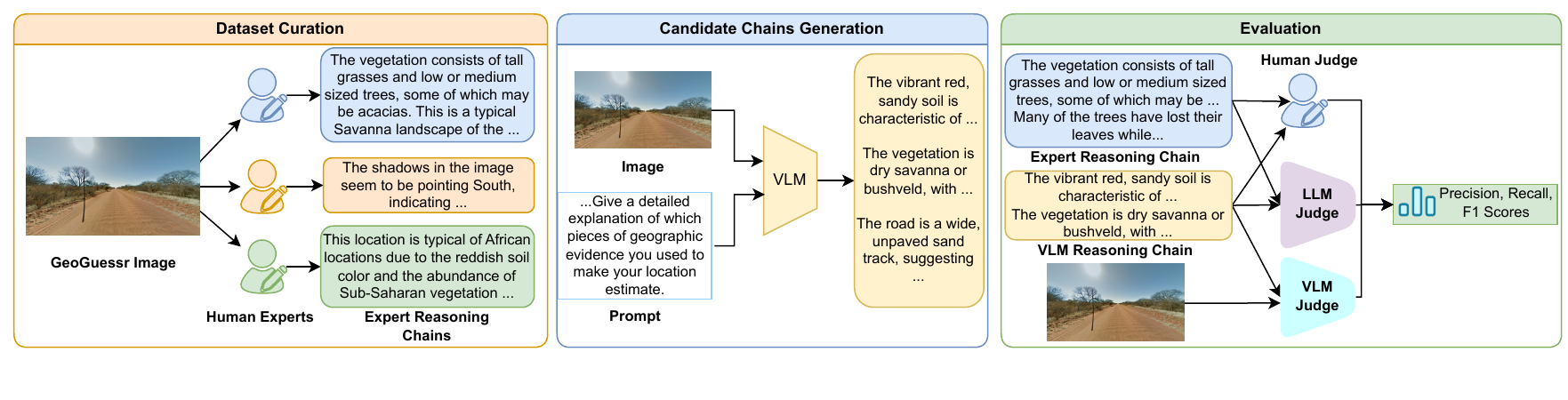}
    \label{fig:summary_fig}
    \vspace{-.4in}
    \caption{GeoRC Benchmark. We curate a dataset of GeoGuessr challenges and associated reasoning chains from three human experts. We then generate reasoning chains from open-weight and proprietary VLMs, and evaluate them through several proposed judging methods.}
\end{figure*}

\begin{abstract}
Vision Language Models (VLMs) are good at recognizing the global location of a photograph -- their geolocation prediction accuracy rivals the best human experts. But many VLMs are startlingly bad at \textit{explaining} which image evidence led to their prediction, even when their location prediction is correct. In this paper, we introduce GeoRC, the first benchmark for geolocation reasoning chains sourced directly from Champion-tier GeoGuessr experts, including the reigning world champion. This benchmark consists of 800 ``ground truth'' reasoning chains across 500 query scenes from GeoGuessr maps, with expert chains addressing hundreds of different discriminative attributes, such as soil properties, architecture, and license plate shapes. We evaluate LLM-as-a-judge and VLM-as-a-judge strategies for scoring VLM-generated reasoning chains against our expert reasoning chains and find that Qwen 3 LLM-as-a-judge correlates best with human-expert scoring. Our benchmark reveals that while large, closed-source VLMs such as Gemini and GPT 5 rival human experts at predicting locations, they still lag behind human experts when it comes to producing auditable reasoning chains. Small open-weight VLMs such as Llama and Qwen catastrophically fail on our benchmark -- they perform only slightly better than a baseline in which an LLM hallucinates a reasoning chain with oracle knowledge of the photo location but \textit{no visual information at all}. We believe the gap between human experts and VLMs on this task points to VLM limitations at extracting fine-grained visual attributes from high resolution images. We open source our benchmark for the community to use.

\end{abstract}

\section{Introduction}

The task of determining the location of a photo has been of interest for more than a century. For example, Frederick Cook claimed to be the first person to climb Denali in 1906 and offered a photograph of the purported summit to support his claim. Cook's claim was discredited by the geolocation of that photo to a different mountain~\cite{washburn1956camera} among other evidence. Beyond photo forensics, global photo geolocation has long been seen as a fun brain teaser, e.g. Cond{\'e} Nast Traveler's ``Where are you?'' competition dating to 1993~\cite{condenast2011where}. 

Today, photo geolocation is still widely relevant as a forensics task, e.g. work by investigative journalists such as Bellingcat and others in the OSINT community, and as a game, e.g. GeoGuessr \cite{geoguessr2013}. In both cases, human experts demonstrate extraordinary skill at using subtle image evidence to determine the location of photographs.

In the last two decades, machine learning approaches have made enormous progress on the global image geolocation task. Starting from im2gps~\cite{im2gps,im2gpsPlus}, methods improved with the introduction of deep learning~\cite{planet_cnn,vo2017revisiting,seo2018cplanet,muller2018geolocation}, then with foundation models such as CLIP~\cite{radford2021learning,clark2023where,cepeda2023geoclip,streetclip,pigeon,g3-guide-book}, and recently with large scale Vision Language Models (VLMs)~\cite{wikitilo,privacy_geolocation,geoChain,navig}. VLMs, with no fine-tuning, appear to roughly match the performance of bespoke geolocation methods.

After two decades of machine learning investigation into geolocation, it is still not clear whether machines or human experts are better at this task. Pigeon~\cite{pigeon} claims that their geolocation method outperforms the best GeoGuessr players. Our experiments suggest that humans still reign supreme. The difference is small and probably depends on the particular experimental setup. It may be the case that the strongest possible geolocation system is a hybrid combination of humans and machines, analogous to the face recognition task where human ``superrecognizers'' have been used as verifiers for machine learning methods~\cite{phillips2018face}.

While the gap in geolocation \textit{accuracy} between human experts and machines may be small, we claim that the gap in \textit{explainability and auditability} is large. When asked to geolocate a photo, human experts can support their decision with specific image evidence related to infrastructure, vegetation, architecture, and a litany of additional attributes that led them to their conclusion. These reasoning chains help establish \textit{trust} -- even a non-expert can verify the presence of the lane markings, writing, or terrain attributes mentioned in the reasoning chain. These explanations also \textit{teach} non-experts to perform the task themselves.

Large generative Vision Language Models can also be asked to \textit{explain their reasoning} at the geolocation task. Generally, they can produce a list of evidence that is qualitatively similar to human expert reasoning chains. However, these VLM reasoning chains often contain \textit{hallucinations}, miss \textit{fine-scale image details}, and exhibit tunnel vision in \textit{rationalizing} the decision made by the VLM.

In this work, we introduce the first purpose-built benchmark of \textit{geolocation reasoning chains}, elicited directly from credentialed \textit{Champion-tier GeoGuessr experts}, including the reigning GeoGuessr world champion, Radu Casapu (RC). We propose a grading scheme to assess how well a candidate reasoning chain agrees with a ``ground truth'' human expert reasoning chain. We deploy an ``LLM-as-a-Judge'' approach to apply our grading scheme to numerous VLMs -- smaller, open weight models such as Gemma \cite{gemma3technicalreport}, Llama \cite{llama3herdmodels}, and Qwen \cite{qwen3technicalreport} as well as large, closed weight models such as Gemini \cite{google2025gemini3} and ChatGPT \cite{openai2025gpt5}. We find a large spread in capability among these models, and we find that the best model (GPT-4.1) still lags behind human experts. 


The contributions of our work include:
\begin{itemize}[noitemsep]
    \item The first purpose-built dataset of human expert geolocation reasoning chains (Section 2)
    \item A grading protocol for humans and machines to evaluate reasoning chain agreement in terms of precision and recall (and summarized with F1 score) (Section 3.1)
    \item An investigation into various LLM-as-a-judge and VLM-as-a-judge methods for assessing reasoning chain quality (Section 3.2)
    \item The first quantification of VLM reasoning chain quality and a characterization of the dominant errors observed -- misattribution, hallucinations, false tool use, axiomatic irrelevance, and missed details (Section 4)
    \item We open source our expert reasoning chains and our best LLM-as-a-judge benchmark for the community to use.
\end{itemize}


\section{Geolocation Reasoning Chains (GeoRC)}

The aim of a geolocation reasoning chain is to detail the thought process of an expert guessing the location from an image. It describes the supporting evidence in an arbitrary order in the form of scene attributes consisting of, but not limited to, infrastructure, architecture, vegetation, climate, geology, terrain, culture, vehicles, and language. In this section, we first characterize a reasoning chain for this task, state its properties, and then explain the details of our dataset.



\subsection{Characterization of GeoRC}
\label{properties-geo-rc}
Generally, we expect that ``good'' geolocation reasoning chains progressively refine an estimate of the location from available evidence, starting from a coarse level (\eg, hemisphere) to a fine level (\eg, city). Reasoning chains typically only contain discriminative scene attributes, rather than being exhaustive. Each attribute should also be associated with a statement of geographic support. For example, ``Short bollards with a vertical reflector and a black `cap' on top are found in Austria and former Yugoslav countries'' cites bollards as a scene attribute alongside relevant geographic regions. Each step in the chain may also represents the confidence level or the degree with which the attribute aids in progressing to the next step in the chain. The ``conclusion'' statement of a reasoning chain consists of the final guess consisting of the country and city or region along with an informal degree of confidence.



\subsection{The GeoRC Dataset}
We curate a dataset, titled GeoRC, consisting of 800 reasoning chains generated by three expert GeoGuessr players.
Our first two experts, Joshua Diao (JD) and Tejas Santanam (TS), are ranked in the Champion Division (top $0.01\%$ of players), while our final expert is the 2025 GeoGuessr World Cup champion and a professional GeoGuessr player, Radu Casapu (RC). Our 800 reasoning chains are generated from 100 GeoGuessr challenges, each comprising 5 unique locations. We task our experts with generating reasoning chains that fit the properties laid out in \Cref{properties-geo-rc}, with each expert writing reasoning chains for a shared set of 150 locations, while the remaining 350 locations are divided among all three experts. See \Cref{app:reasoning_chain_guidelines} for detailed instructions provided to our GeoGuessr players.

\begin{figure*}[h!]
    \centering
    \includegraphics[width=\textwidth]{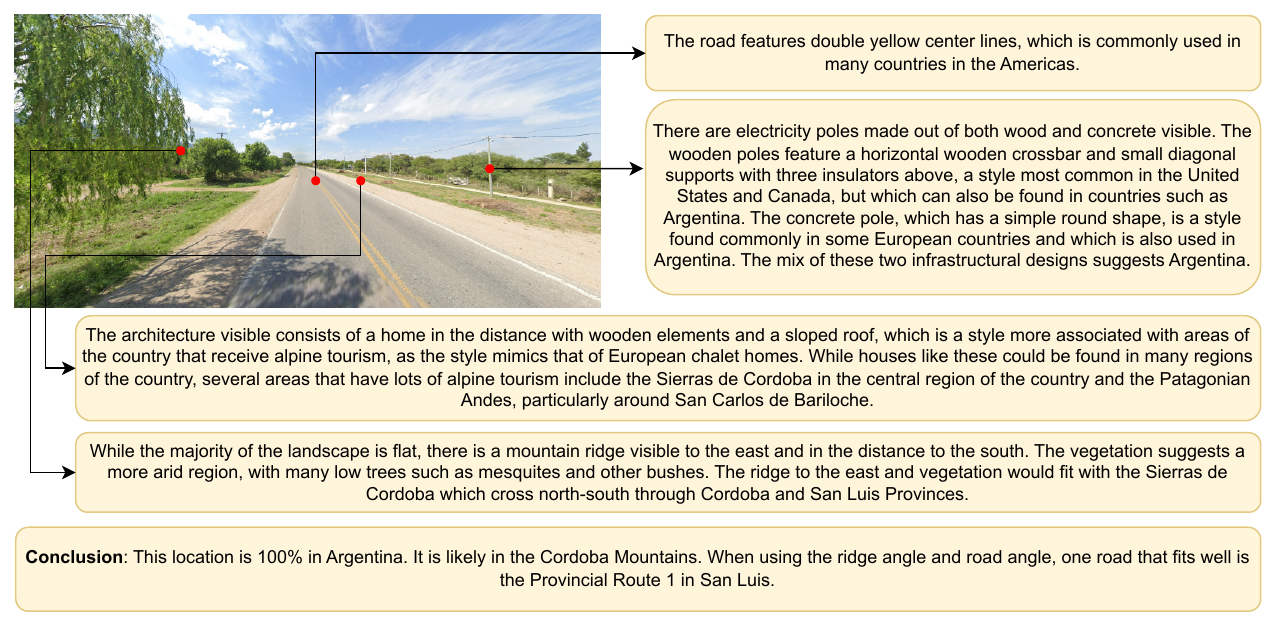}
    \captionsetup{skip=1pt}
    \caption{\textbf{Example of Expert Reasoning Chain.} Experts typically note discriminative visual features in a coarse-to-fine manner before concluding with a country guess. Expert chains are usually non-exhaustive, only requiring a small number of keypoints to localize the image to a country or region.}
    \label{expert_chains}
\end{figure*}

\subsection{Geographic Scene Attribute Categories}

\begin{figure}[h!]
    \centering
    \includegraphics[width=\linewidth]{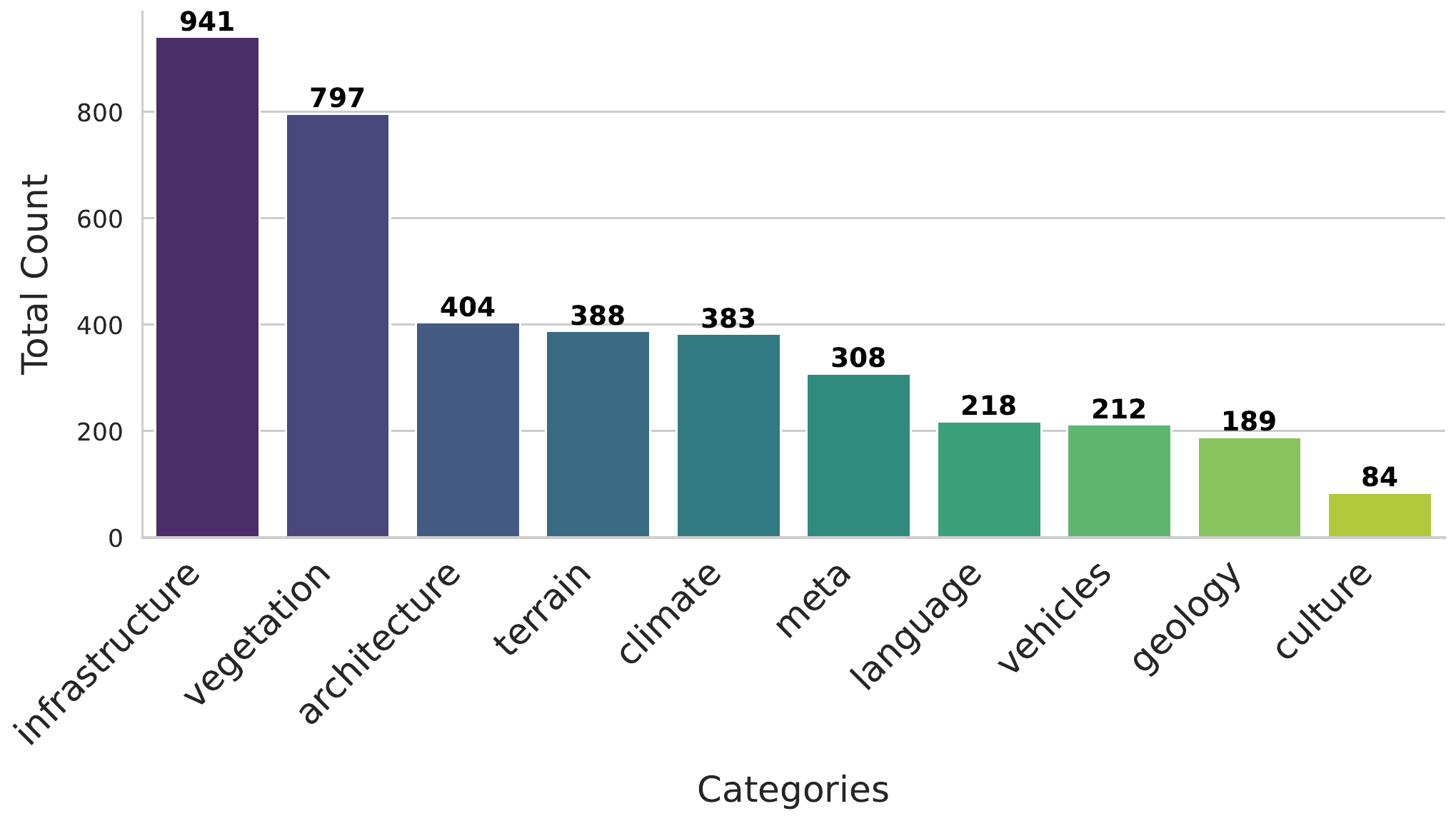}
    \caption{Categories of Geographic Scene Attributes cited by our GeoGuessr experts. Infrastructure and vegetation are the top cited scene attributes. }
    \label{fig:categories_hist}
\end{figure}

The non-exhaustive reasoning chains in the GeoRC dataset cite a wide variety of geographic scene attributes, which can be broadly categorized. The most cited categories are the infrastructure visible (\eg poles, bollards), followed by vegetation and architecture. One unique category is ``meta information,'' which includes GeoGuessr and Street View-specific cues such as the Street View car, camera quality, and map coverage. The language category typically refers to signs with visible language. The least cited category happens to be the culture that is unique to a specific region or country. Additional categories include terrain, climate, vehicles, and geology. After having our experts write reasoning chains, we ask an LLM to categorize and label each point in a reasoning chain into a maximum of three categories. \Cref{fig:categories_hist} shows the count of these categories across our 800 expert chains. 

\subsection{Geographical Distribution}

\begin{figure}[h!]
    \centering
    \includegraphics[width=\linewidth]{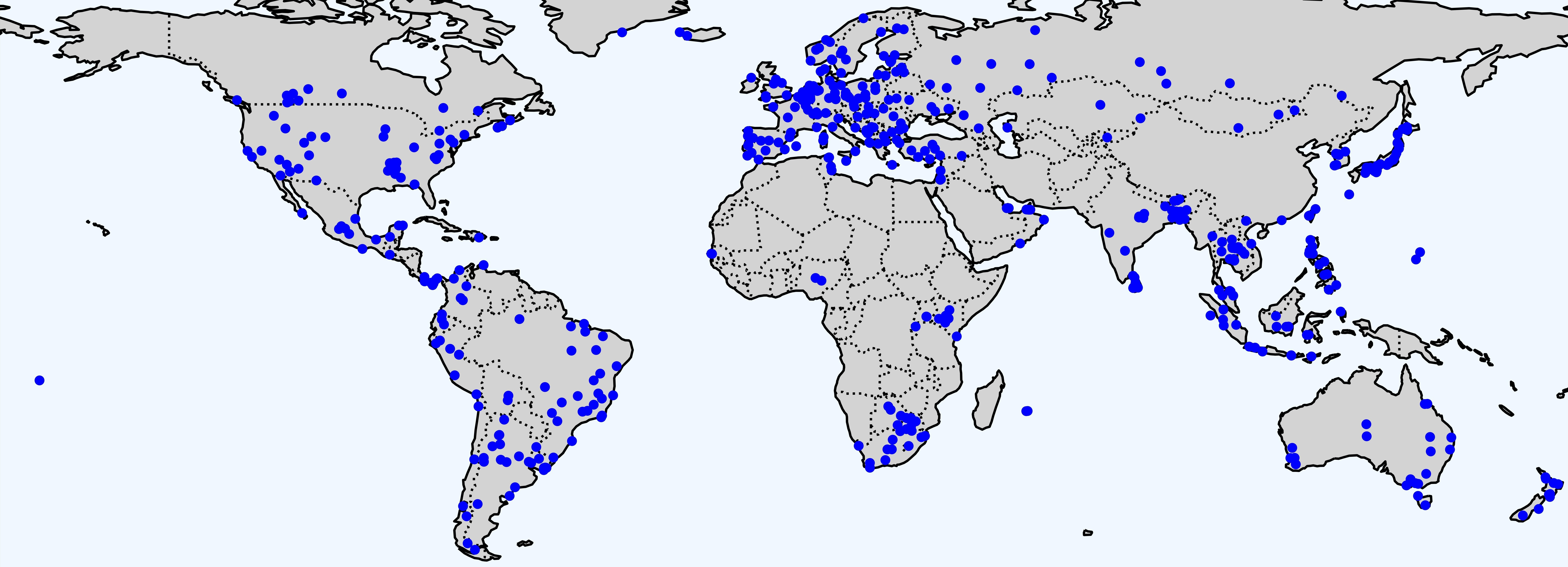}
    \caption{Geographical Distribution of GeoRC Dataset that is drawn from popular GeoGuessr world maps inherently conditioned upon Google Street View coverage.}
    \label{geo_dist}
\end{figure}

Distribution of chosen locations in our dataset are derived from popular maps on GeoGuessr, such as GeoGuessr Saturday, An Arbitrary World, and An Arbitrary Urban World, all of which are inherently conditioned upon the distribution of official Google Street View coverage. These maps are curated through a random sampling of latitude and longitude, followed by selecting the nearest street view image to the randomly sampled location coordinates \cite{vercel_map}. Sampling of the maps is also parameterized, such as ``urban'' for the Arbitrary Urban World map to indicate urban city landscapes \cite{vali}. The geographical distribution of our dataset is represented in Figure \ref{geo_dist}.

\section{Evaluating Reasoning Chains}

We aim to develop an automated method for assessing candidate reasoning chains against human expert chains. In this section, we describe the process of human grading, which we use for calibration, and propose three methods for grading candidate reasoning chains. The first two approaches utilize an LLM to measure the relationship between each point in the candidate chain and points in the ground truth chain. The third approach utilizes an open source VLM in addition to an LLM to score the candidates. In-depth algorithm pseudocode can be found in \Cref{app:pseudocode}.

\subsection{Human Grading}

We asked our expert GeoGuessr players to evaluate 150 candidate reasoning chains by adopting the one-to-all bipartite strategy. In this strategy, each candidate step is compared to all steps in the reference chain. The overall score is computed by averaging across all points. Each comparison direction gives two measures, precision and recall, which are then used to calculate the F1 score. An illustration of the approach on a candidate reasoning chain is shown in \Cref{human_scoring_ex}. To avoid bias, graders were assigned reasoning chains they did not write. Exact grading guidelines can be found in \Cref{app:human_grading_guidelines}. 

\subsection{Approach 1: One-to-all LLM-as-a-judge Evaluation}
\label{approach_1}
In this approach, we utilize only an LLM judge to determine the degree of similarity of each point in the candidate reasoning chain with the points in the ground truth reasoning chain. In our prompt to score a given candidate point against the complete ground truth chain, the LLM judge is provided the context, a set of rules to adhere to, the complete ground truth chain, and a single candidate point and is asked to respond with a similarity score out of 100. We request the LLM judge over each candidate point and then compute an average score for the complete chain. This results in the precision score for our grading. Similarly, when iterating through each ground truth point and the complete candidate chain, we obtain the recall score for our grading. Using the precision and recall together, we compute the F1 score. The algorithm for this approach is shown in \Cref{one_vs_all_llm} with chain 1 being the candidate chain and chain 2 the ground truth chain for precision and vice versa for recall. We had also experimented with simpler strategies, such as both the chains scored in their entirety, but these deviated from the human grading expectations.

\begin{figure*}[h!]
    \centering
    \includegraphics[width=0.95\textwidth]{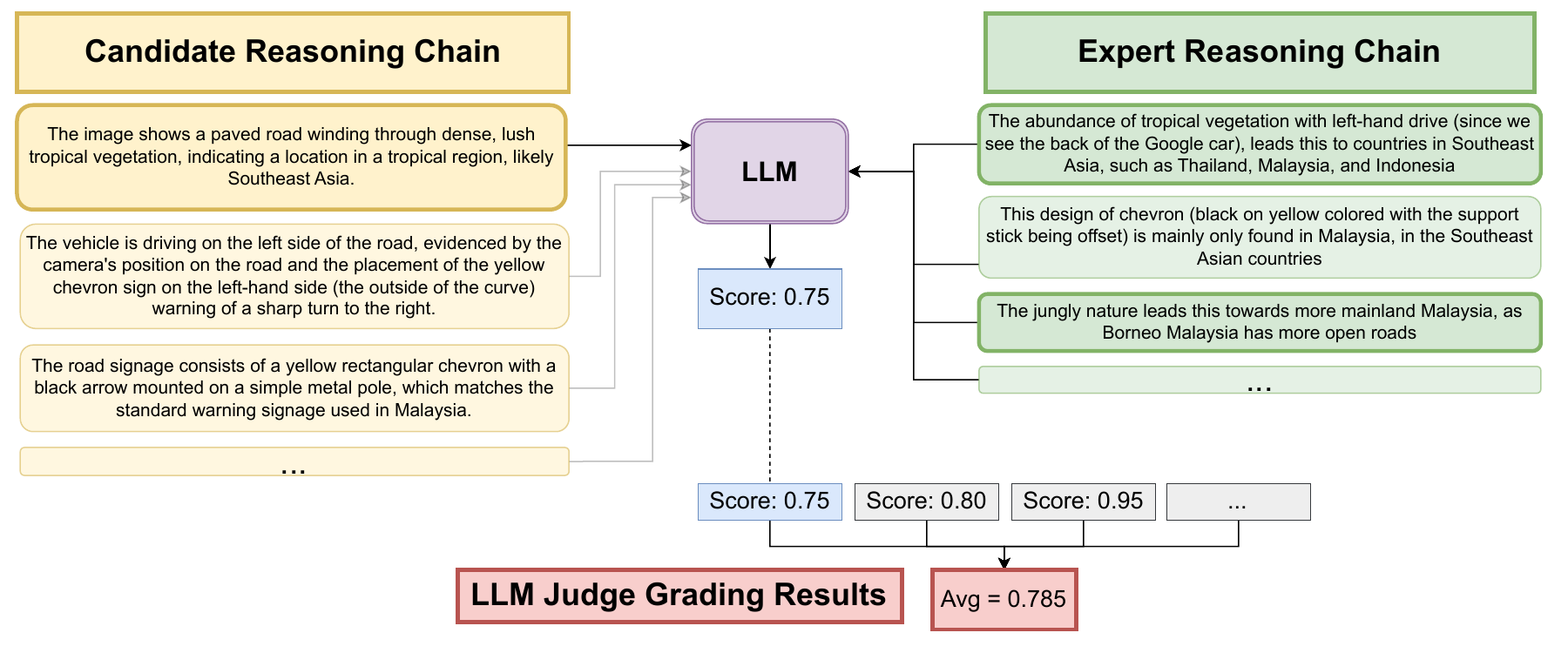}
    \caption{One-To-All LLM-as-a-judge Evaluation. One step in the candidate chain is compared to all steps in the expert reasoning chain to compute an F1 score by an LLM judge. An average across all steps in the candidate chain results into the overall score for this candidate.}
    \label{fig:kp_scoring}
\end{figure*}

\subsection{Approach 2: Key Points Guided LLM-as-a-judge Evaluation}

To further enhance the reliability, consistency, and to mitigate biases with LLM-as-a-judge (\cite{llm_judge}, \cite{score_range_bias_llm_judge}), we propose this key points guided approach. The complex scoring task is broken down into granular, easier tasks for the LLM judge. Key points, instead of individual bullet points, are input to the LLM. A key point (appendix section C \cite{molmo}) is a clause in natural language that summarizes the main idea of the step in the chain by extracting words from the step. A single step is represented by one to three key points, with each key point weighted in an unbiased manner. Each key point is then transformed into an embedding space by a sentence transformer \cite{sentence_transformer}. Cosine similarity is computed between each pair of their vectors and is thresholded. Thresholding hyperparameters are tuned to align with human grading. The score is also normalized. See \Cref{key_points_llm} for a pseudo-code implementation and \Cref{fig:kp_scoring}.
\vspace{-4pt}
\subsection{Approach 3: VLM-as-a-judge Evaluation}
\vspace{-3pt}
We hypothesize that by supplying the original image to the judge, the judge may better identify hallucinations. Therefore, we develop a scoring approach that utilizes a VLM to compute the correctness of the candidate points. We prompt the VLM judge to output the number of statements that are corroborated by the image, thus constructing a correctness score for the candidate chain. Subsequently, we utilize the same strategy as that adopted in \Cref{approach_1} to compute the precision and recall of the chain. In order to reduce the computational cost associated with inferencing both a VLM and an LLM, we make a single request to the LLM judge for measuring the precision and recall instead of its granular one-to-all counterpart. \Cref{vlm_judge} shows the pseudo-code implementation.

\section{GeoRC Benchmark}

\subsection{Experimental Setup}

We execute our experiments on two Nvidia A40 GPUs. Qwen3-4B-Instruct-2507 \cite{qwen3technicalreport} and Qwen2.5-VL-72B-Instruct \cite{qwen25vltechnicalreport} are chosen to be the LLM and VLM judges for our evaluation methods, respectively.

\subsection{Baseline Methods}
We introduce three baseline methods of candidate reasoning chains that serve as reference points for both extremes of the scoring spectrum.
    \paragraph{Hallucinated Reasoning Chains.} An LLM is supplied context of the country and city of where the image was captured, but \textit{not the image itself}. It is then prompted to generate a geographical reasoning chain following the list of common categories of scene attributes to consider. The resulting chains consist of hallucinated scene attributes that may not be present in the image. Evaluation scores for this candidate should be relatively low.
    \paragraph{Random Reasoning Chains.} From the human expert reasoning chains, we randomly choose a chain from an entirely different location than a given reference chain. As the scene attributes are highly unlikely to overlap between random location pairs, evaluation scores for this candidate should be near-zero.
    \paragraph{Paraphrased Reasoning Chains.} An LLM is given the best expert's reasoning chain as reference and is prompted to generate a paraphrased version of the chain. Evaluation scores for this candidate should be relatively high.

\subsection{Evaluation of Judging Methods}


We compare our automated evaluation methods with human grading on a subset of 225 pairs consisting of 75 locations across 3 types of candidates and report the Mean Absolute Error for each candidate type and Pearson, Spearman and Kendall Correlation Coefficients for all candidates in \Cref{tab:mae_results}. We find that the One-to-all approach aligns best with human grading on this subset of candidates. The Inter Annotator Agreement metrics observed across the experts for the same subset of candidate scores is shown in \Cref{tab:ICC}. 

\begin{table}[!ht]
\centering
\label{tab:mae_results}
\begin{adjustbox}{max width=\linewidth}
\begin{tabular}{lccc}
\toprule
\textbf{Experiment} & \textbf{One-To-All} & \textbf{KeyPoint} & \textbf{VLM-based} \\
\midrule
Expert vs Best Expert & \textbf{12.06} & 12.72 & 16.97 \\
Hallucinated vs Best Expert & \textbf{11.58} & 13.98 & 15.65 \\
VLM vs Best Expert & \textbf{12.54} & 12.79 & 24.00 \\
\midrule
Pearson Correlation Coefficient & \textbf{0.6893} & 0.6766 & 0.4951 \\
\bottomrule
\end{tabular}
\end{adjustbox}
\caption{Evaluation of Judging Methods using Mean Absolute Error and Correlation Coefficients}
\end{table}

\subsection{Benchmark Results}

Table \ref{main_results_table} shows the precision, recall, and F1 metrics from scoring multiple candidate reasoning chains with the One-to-all scoring approach. We also present the geolocation accuracy, which is the percentage of correct country predictions by each candidate as measured by an LLM judge against the ground truth location country in ISO format. 

Human expert reasoning chains achieve an average F1 score of 56, by comparing distinct expert candidate \& reference chains paired as 1 \& 2, 2 \& 3, and 1 \& 3 over 150 locations each. As mentioned in \Cref{properties-geo-rc}, reasoning chains are non-exhaustive and therefore, a different variety of scene attributes is referenced by each expert. Baseline candidates achieve scores as expected across the spectrum, with random hallucinated candidates scoring lowest, hallucinated candidates scoring 18.13, and paraphrased candidates scoring the highest. Interestingly, open-source VLMs such as Llama-3.2 and Qwen-3 score close to the hallucinated baseline. This implies these VLMs glean the least amount of scene information from the images, so much so that they perform about the same as when the image is not supplied at all. Qwen-2.5VL is the best performing in this category. However, it lags behind human experts by about 20 points. Furthermore, another interesting observation is that for Qwen2.5, recall is higher than precision. This is because its chains mostly contain irrelevant non-discriminative attributes that do not aid reasoning. 

Proprietary VLMs perform significantly better than open-weight VLMs. The reasoning chains they generate are of superior quality in terms of conciseness and practicality. GPT-4.1 is the best performing VLM. However, there exists a significant \textbf{12 point gap} between its score and the average F1 score of the human expert chains.

\begin{figure}[h!]
  \centering
  \includegraphics[width=\linewidth]{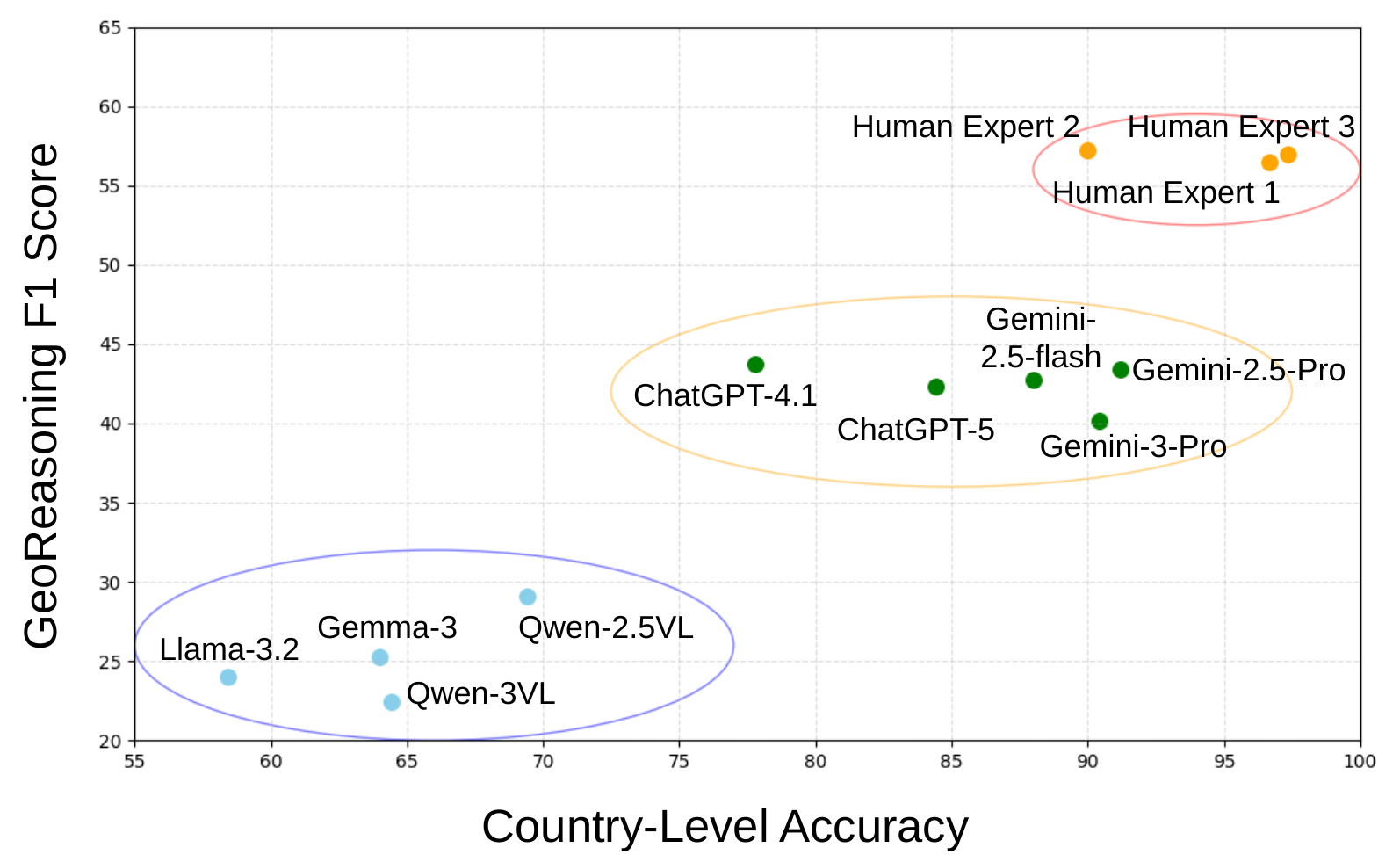}
  \caption{Distinct groupings observed for candidates in the graph for their F1 scores and Country-level accuracy }
  \label{accuracy_f1_plot}
\end{figure}

When scores are graphed against country-level geolocation accuracy, distinct clusters are observed across these categories of candidates, namely, experts, open-weight VLMs, and proprietary VLMs, as shown in Figure \ref{accuracy_f1_plot}. The open-weight VLMs perform the worst in both F1 scores and accuracy. Proprietary models perform better, but still fall behind our human experts. 

\setheatlimits{20}{50}
\begin{table}[!t]

\centering

\begin{adjustbox}{max width=\linewidth}
\begin{tabular}{l c c c c c}

\toprule

& \multicolumn{4}{c}{\textbf{Complete Benchmark}}
& \textbf{Country Acc.} \\
\cmidrule(lr){2-5}  \cmidrule(lr){6-6}

\textbf{Candidate} &  \textbf{No.}
& \textbf{Precision} & \textbf{Recall} & \textbf{F1}
& \textbf{\%} \\
\midrule

\multicolumn{6}{c}{\textbf{Human Expert}} \\
\midrule
Human Expert \#1 - JD &  -- & -- & -- & -- &  \setheatlimits{55}{95} \heat{96.67} \setheatlimits{20}{50}\\

Human Expert \#2 - TS &  -- & -- & -- & -- & \setheatlimits{55}{95} \heat{90.00} \setheatlimits{20}{50}\\
Human Expert \#3 - RC (Best Expert) &  -- & -- & -- & -- & \setheatlimits{55}{95} \textbf{\heat{97.33}} \setheatlimits{20}{50}\\
Human Expert Average &  450  & \textbf{\heat{55.06}} & \textbf{\heat{65.41}} & \textbf{\heat{56.69}} & -- \\

\midrule
\multicolumn{6}{c}{\textbf{Baseline}} \\
\midrule
Random & 800 & \heat{2.75} & \heat{2.52} & \heat{1.90} & -- \\
Hallucinated & 800 & \heat{27.91} & \heat{20.12} & \heat{18.13} & -- \\
Paraphrased &800 &\heat{ 97.51 }&\heat{ 98.47 }&\heat{ 97.93 }&--\\

\midrule
\multicolumn{6}{c}{\textbf{Open-weight VLMs}} \\
\midrule
Llama-3.2-11B-Vision-Instruct \shortcite{llama3herdmodels} &800&\heat{ 22.26 }&\heat{ 36.86 }&\heat{ 24.00 }&\setheatlimits{55}{95}\heat{ 58.40}\setheatlimits{20}{50}\\
Qwen2.5-VL-7B-Instruct \shortcite{qwen25vltechnicalreport} &800&\heat{ 23.95}&\heat{ 48.26 }&\heat{ 29.09 }&\setheatlimits{55}{95}\heat{ 69.40}\setheatlimits{20}{50}\\
Qwen3-VL-8B-Instruct \shortcite{qwen3vltechnicalreport} &800&\heat{ 26.78 }&\heat{ 25.15 }&\heat{ 22.41 }&\setheatlimits{55}{95}\heat{ 64.40}\setheatlimits{20}{50}\\
Gemma-3-12b-it \shortcite{gemma3technicalreport} &800&\heat{ 23.55 }&\heat{ 35.55 }&\heat{ 25.23 }&\setheatlimits{55}{95}\heat{ 64.00}\setheatlimits{20}{50}\\

\midrule
\multicolumn{6}{c}{\textbf{Proprietary VLMs}} \\
\midrule
GPT-5 \shortcite{openai2025gpt5}&800&\heat{ 49.68 }&\heat{ 42.14 }&\heat{ 42.31 }&\setheatlimits{55}{95}\heat{ 88.40} \setheatlimits{20}{50}\\
GPT-4.1 \shortcite{openai2025gpt41}&800&\heat{ 46.19 }&\heat{ 47.58 }&\heat{ 43.74 }&\setheatlimits{55}{95}\heat{ 77.80}\setheatlimits{20}{50}\\
Gemini-3-Pro \shortcite{google2025gemini3}&800&\heat{ 52.17 }&\heat{ 38.17 }&\heat{ 40.19 }&\setheatlimits{55}{95}\heat{ 90.40}\setheatlimits{20}{50}\\
Gemini-2.5-Pro \shortcite{gemini25}&800&\heat{ 49.59 }&\heat{ 44.10 }&\heat{ 43.42 }&\setheatlimits{55}{95}\heat{ 91.20}\setheatlimits{20}{50}\\
Gemini-2.5-Flash \shortcite{gemini25}&800&\heat{ 48.15 }&\heat{ 44.30 }&\heat{ 42.76 }&\setheatlimits{55}{95}\heat{ 88.00}\setheatlimits{20}{50}\\
\bottomrule
\end{tabular}
\end{adjustbox}
\caption{Evaluation of Candidate Reasoning Chains using the One-To-All Scoring Method}
\label{main_results_table}
\end{table}

\subsection{Qualitative Results}

\begin{figure*}[h!]
    \centering
    \includegraphics[width=\textwidth]{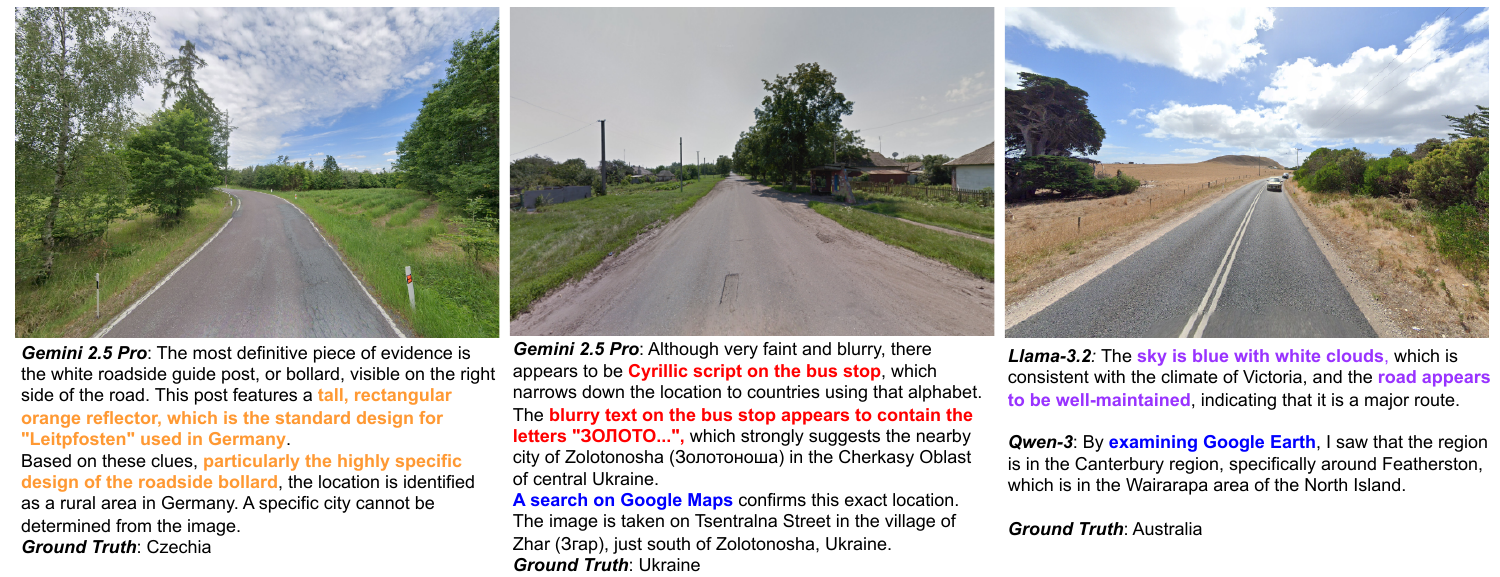}
    \caption{Text highlighted in \textbf{\textcolor{orange}{orange}} shows geographic misattribution where the scene attribute despite being in multiple countries is associated with a specific country. Text highlighted in \textbf{\textcolor{red}{red}} shows a hallucination where the referred scene attribute is absent and is not corroborated by the image. Text highlighted in \textbf{\textcolor{blue}{blue}} shows False Tool Use. Text highlighted in \textbf{\textcolor{MyPurple}{purple}} depicts an axiomatic irrelevance which is an obvious statement made about a geolocation attribute that is not contributing to the reasoning}
    \label{fig:common_error_scenarios}
\end{figure*}

Throughout the low scoring geolocation reasoning chains candidates generated by both closed source and open source VLMs, we observed that the causes for the low score are attributed to first, the textual generation and two, the visual processing of the input image. Firstly, the text generated by the VLM candidates jumps straight to the conclusion while the subsequent steps in the chain serve as rationalizations for the guessed location. This tendency to consider scene attributes as rationalizations \cite{post_hoc_rationalization} rather than steps in a chain of reasoning causes VLMs to falter and produce text that is quite error prone. We classify these errors into four categories.  \cref{fig:common_error_scenarios} and \cref{common_error_scenarios_appendix} shows images and excerpts with each category color-coded.
    \paragraph{Geographic Misattribution.} VLMs tend to hastily draw conclusions from geographic attributes, especially roadside infrastructure, landscapes, and housing architecture, and misattribute it as discriminating against a single country. Conversely, human experts consider various geolocation possibilities for a scene attribute and then progressively narrow down the location based on subsequently observed scene attributes.
    \paragraph{Hallucination.} To rationalize their conclusions, VLMs concoct information that is not corroborated by the input image. Even the best performing closed source VLMs hallucinate about the language, architecture, and infrastructure. The degree of hallucinations is often severe, to the extent that buildings are labeled as specific industrial complexes from a different country, and road signs are said to be present with text in a specific language.
    \paragraph{False Tool Use.} The text generated by both open and closed-source VLMs cites tools such as Google Maps, Google Street View, and Google Earth, despite the fact that no tools are available to the VLMs. 
    \paragraph{Axiomatic Irrelevance.} This category of failures depicts premises stating facts that are almost always true, but are largely nondiscriminative. Features such as ``sky is blue with white clouds'' and ``well-maintained road'' are vague as they are true for a wide variety of locations, and hence cannot contribute to guessing the location.

\begin{figure}
    \centering
    \includegraphics[width=\linewidth]{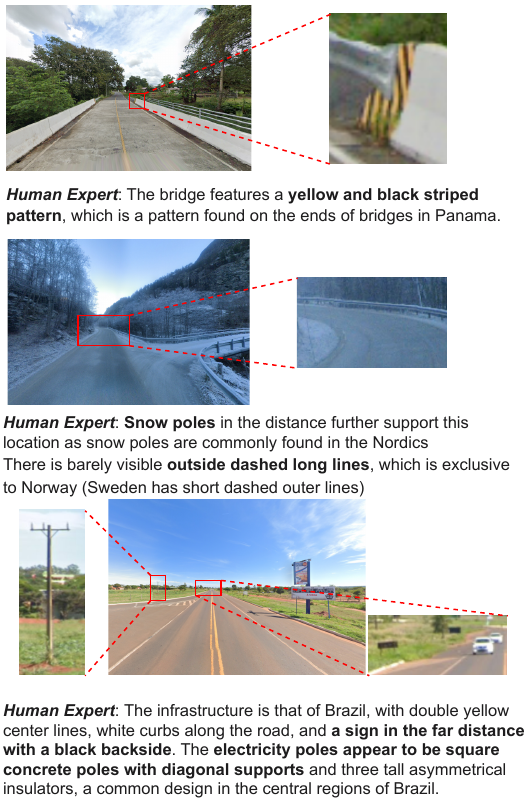}
    \caption{Examples of images where scene attributes with relatively small pixel size are overlooked by the candidate VLMs}
    \label{pixel_limitations}
\end{figure}

Another cause for the low candidate VLMs scores might be lossy input image encodings.  Scene attributes utilized by our geolocation experts are often quite small, hence potentially being lost should any downsampling of the input occur. In contrast, most scene attributes cited by the VLM candidates occupy a large number of pixels in the image, overlooking these small, but insightful pieces of evidence. For example, as shown in \Cref{pixel_limitations}, candidate VLMs, including the highest performing proprietary models, fail to mention the yellow and black striped pattern on the ends of the bridge, snow poles along the roadside at a distance, dashed lines on the road, distant road signs and utility poles that are visible to the human expert when they inspect the image more closely. See appendix \ref{ablation_pixel_space} for concrete examples.

The above characterizations reveal that both the visual encoder and text generation modules suffer from limitations on the geolocation task.


\section{Related Work}



Several prior works have explored both the generation of reasoning chains and the evaluation of VLMs on the geolocation task by analyzing individual clues. A summary of the comparison to related work is shown in \Cref{related_work_comparison}. GeoReasoner \cite{geoReasoner} is a large Vision Language Model that is trained predominantly on Google Street View images and textual clues from geolocalization games, including GeoGuessr \cite{geoguessr2013} and Tuxun \cite{tuxunfun}. However, their reasoning is learned by the model from unstructured textual clues that are limited to facts about locations posted by the communities within the games. Furthermore, the image-text pairs are presented in a fixed question and answer format, for which the answers are filtered and generated by a different model, not by a human expert. In contrast, we propose human-written structured reasoning chains that are non-exhaustive, unconstrained, and structured, identifying multiple pieces of evidence for reasoning about an image's location. Another work, GeoChain \cite{geoChain}, is similar as its data is not human written, constrained, and exhaustive, as it evaluates the reasoning capabilities using a fixed set of 21 questions over four categories shared across the entire dataset. GeoChain additionally utilizes a semantic segmentation model to derive answers to a fixed set of questions about the scene, limiting reasoning evaluation to the categories identified by the model. We distinguish ourselves by not confining our approach to fixed categories, utilizing chains written by expert GeoGuessr players that encapsulate unique and creative attributes, devoid of any semantic analyzer. 

Other works have explored utilizing open-ended reasoning chains to improve or evaluate VLM geolocation capabilities, but primarily differ from our work in data sourcing and evaluation rigor. GRE30K \cite{gre30k} and MP16-Reason \cite{mp16Reason} source synthetic reasoning chains by prompting VLMs with data from the MP16 dataset \cite{larson2017benchmarking}, utilizing these reasoning traces as supervision for training geolocation reasoning models, and in the former case, evaluation via GPT-4o, RefCLIPScore, and BERTScore. Other approaches source human reasoning chains ``in-the-wild'' from social media: GeoExplain \cite{geoExplain} builds a dataset by scraping Reddit comments, while NAVIg \cite{navig} scrapes YouTube video transcripts. Due to an additional focus on model supervision, these reasoning chains are naturally unstructured and are evaluated with uncalibrated metrics such as ROUGE, BLEU, and BERTScore.

The closest work to ours is GeoComp \cite{geoComp}, which sources human reasoning by having three ``gaming enthusiasts'' collaborate on a consensus chain per image. These chains are then used for evaluating VLM outputs via an uncalibrated GPTScore with GPT-4o. However, GeoComp does not establish a fixed annotation protocol to be shared by all annotators, which results in some of their chains containing non-discriminative cosmetic remarks, rather than exclusively containing features that were deemed important by an expert. Our work addresses these limitations by developing a dataset with three experts under a strict, standardized protocol, followed by evaluation with a calibrated judging framework.

Outside of geolocation reasoning, WikiTiLo \cite{wikitilo} also adopts a similar question answering based strategy restricted to forming answers only for guessing the country, city, and time of an image. It does not focus on scene attributes that are crucial for contributing to reasoning for geolocation. Their F1 score evaluation is contingent upon only the answers to these 3 questions. 


\definecolor{tablegray}{gray}{0.95} 

\begin{table}[ht] 
    \centering
    \scriptsize 
    \renewcommand{\arraystretch}{1.2}
    \setlength{\tabcolsep}{2pt} 
    \begin{tabularx}{0.48\textwidth}{
        >{\hsize=0.9\hsize\raggedright\arraybackslash}X 
        >{\hsize=0.3\hsize\centering\arraybackslash}X   
        >{\hsize=0.3\hsize\centering\arraybackslash}X   
        >{\hsize=1.1\hsize\raggedright\arraybackslash}X 
        >{\hsize=1.1\hsize\raggedright\arraybackslash}X 
        >{\hsize=0.5\hsize\centering\arraybackslash}X   
        >{\hsize=0.5\hsize\centering\arraybackslash}X   
    }
        \toprule
        \rowcolor{gray!15} 
        & \multicolumn{2}{c}{\textbf{Focus}} & & & & \\
        \rowcolor{gray!15} 
        \textbf{Dataset} & \textbf{B.} & \textbf{M.} & \textbf{Source} & \textbf{Eval} & \textbf{Open Judge} & \textbf{Human} \\ 
        \midrule
        GRE30K \shortcite{gre30k} & \checkmark & \checkmark & GPT-o3 & Recall, BERT & $\times$ & $\times$ \\ 
        \rowcolor{tablegray} 
        MP16-Reason \shortcite{mp16Reason} & $\times$ & \checkmark & 3 VLMs & N/A & N/A & N/A \\ 
        GeoExplain \shortcite{geoExplain} & \checkmark & \checkmark & Reddit & ROUGE, BERT & \checkmark & $\times$ \\ 
        \rowcolor{tablegray} 
        NAVIg \shortcite{navig} & $\times$ & \checkmark & YouTube & ROUGE & \checkmark & $\times$ \\ 
        GeoComp \shortcite{geoComp} & \checkmark & \checkmark & Gaming Enthusiasts & GPTScore & $\times$ & $\times$ \\ 
        \rowcolor{tablegray} 
        GeoReasoner \shortcite{geoReasoner} & $\times$ & \checkmark & Textual clues & Fixed Qs & $\times$ & $\times$ \\ 
        GeoChain \shortcite{geoChain} & \checkmark & $\times$ & Semantic Model & 21 Qs & $\times$ & $\times$ \\ 
        \rowcolor{blue!5} 
        \textbf{GeoRC} (Ours) & \checkmark & $\times$ & \textbf{GeoGuessr Experts} & \textbf{LLM-Judge} & \checkmark & \checkmark \\ 
        \bottomrule
    \end{tabularx}
    \caption{Comparison to Related Work based on focus (benchmarking reasoning, model improvement), source, evaluation method, open source judge, and human validation of evaluation method. Our work utilizes human-written reasoning chains, focuses exclusively on benchmarking, and employs human grading for score calibration.}
    \label{related_work_comparison}
\end{table}
\section{Conclusion}

In this work, we introduced GeoRC, a benchmark consisting of human-expert geolocation reasoning chains. By collecting reasoning chains from three experts following a shared annotation protocol, utilizing human-grading for calibrating automated evaluation methods, and measuring human-to-human scoring, we reveal a 12 point gap in the geolocation reasoning ability of frontier VLMs and human experts. Our analysis of VLM failure modes finds that this gap stems from the widespread prevalence of hallucinations, geographical misattributions, red herrings, axiomatic irrelevances, and the inability to recognize low-pixel-count attributes. Future work to enhance VLM reasoning must address these drawbacks by improving the vision encoding modules to focus on much finer image scene attributes and discourage post-hoc rationalization through reward signals during the training process. Generation of explainable and auditable reasoning traces will bring us multiple steps closer towards understanding the interactions of the text and vision modalities in VLMs and develop better VLMs for more complex cognitive tasks. 

\section{Limitations}
First, we note that our expert reasoning chains are non-exhaustive. Accordingly, it may be possible (though unlikely) that fully disjoint sets of \textit{true} statements could be used to arrive at the same country guess. Additionally, our experts occasionally used additional context, such as the compass, to determine the hemisphere or to adopt certain map alignment strategies that were not supplied to the VLMs. Another limitation is that the language for our reasoning chains is only in English.
We also note that our experiments rely on a fixed prompt between all VLMs tested. We acknowledge that each VLM could perform more competitively with humans with specific prompt tuning. Furthermore, our compute for the open-source VLMs was insufficient to run the largest and most capable open-weight models, such as Gemma 3-27B \cite{gemma3technicalreport} and Qwen3-VL-235B \cite{qwen3vltechnicalreport}. 

Our work at present does not have expert-annotated visual grounding input for the image, which would be beneficial for the research community to analyze the vision components of the VLMs. Another aspect that could be a value addition would be for investigating the post-hoc rationalization wherein reasoning chains could detail the tacit knowledge of our experts, their introspection, and deliberation methods. Both of these are exciting directions for future work. 

\section{Ethical Considerations}

Generative AI tools were used to fix grammar and debug code. 

Since our reasoning benchmark seeks to evaluate and eventually improve the task of geolocation reasoning, it poses privacy risks~\cite{privacy_geolocation} if it improves the state of the art in image geolocation. On the other hand, better explanations as to \textit{why} a photo can be geolocated would also allow people to \textit{protect} their privacy by obscuring those details.

Unlike datasets that rely on crowdsourced labor, this work did not involve external annotators. The geolocation reasoning chains were authored directly by three domain experts who are members of the research team and co-authors of this paper. Their intellectual contributions extend throughout the project, including defining the research problem, establishing the protocol for creating a reasoning chain, and developing the experimental codebase. As such, standard ethical reporting regarding annotator recruitment and piece-rate compensation does not apply.

We license the GeoRC reasoning chains under the CC BY license and the GeoRC benchmarking code under the Apache license. We also share code to reconstitute and download the query images in a manner that complies with GeoGuessr and Google Street View’s terms of service.

\bibliography{custom}

\newpage
\appendix

\section{Appendix}
\label{sec:appendix}
\subsection{Human Reasoning Chain Guidelines}
\label{app:reasoning_chain_guidelines}

\begin{figure*}[h]
    \centering
    \begin{quotebox}
        \begin{itemize}
  \item A \emph{reasoning chain} should follow the actual thought process of the expert. There is no preferred order (e.g.\ vegetation first, then architecture).

  \item Reasoning chains can be arbitrarily long. When substantial supporting evidence is present, it should be mentioned.

  \item Typical statements identify an attribute and then describe its approximate geographic support. For example:
  \begin{quote}
    ``Short bollards with a vertical reflector and a black `cap' on top are found in Austria and former Yugoslav countries.''
  \end{quote}
  This is preferred to categorical statements such as ``this is an Austria bollard,'' because visual descriptions are easier to verify and inspire more trust.

  \item It is not necessary to explain every attribute in exhaustive detail. However, including a few distinguishing features can be helpful. For instance, instead of stating ``This is a Paraná pine,'' it is better to describe enough features that a non-expert could identify it among other plants in the image.

  \item Not every attribute requires precise geographic support. Broad descriptors such as ``arid,'' ``temperate,'' or ``tropical'' are useful despite being geographically widespread.

  \item Statements that limit the predictive power of an attribute are valuable. For example:
  \begin{quote}
    ``Thai architecture is fairly uniform throughout the country with only small variations, so there is not enough architectural information within the image to give any confidence on the region.''
  \end{quote}

  \item Because reasoning chains build on earlier points, statements may assume that the region has already been narrowed down. For example:
  \begin{quote}
    ``The soil is a dark red-orange color which is indicative of Western Australia, but can sometimes be found in other parts of the country.''
  \end{quote}

  \item Evidence varies in diagnostic strength. Some cues are weakly diagnostic (e.g.\ ``arid climate''), while others are highly diagnostic (e.g.\ ``concrete roads and electrical poles with a single insulator,'' which implies the Philippines). Statements should reflect this uncertainty, such as:
  \begin{quote}
    ``The road infrastructure is in poor condition, but this only slightly helps to narrow down the location.''
  \end{quote}

  \item Avoid referring to yourself. Instead of ``The grass reminds me of X,'' write ``The grass is typical of X'' or ``Similar scenes are found in X.'' Instead of ``I can’t tell if this is a Canadian flag,'' write ``There is a flag, but it is difficult to distinguish.'' An exception may be made in the conclusion, where first-person phrasing can feel more natural.
\end{itemize}
    \end{quotebox}
    \caption{Reasoning Chain Guidelines}
    \label{box:reasoning_guidelines}
\end{figure*}

\begin{figure*}[h]
    \centering
    \begin{quotebox}
        \begin{itemize}
  \item The conclusion should clearly characterize the uncertainty of the prediction. It should describe the likely geographic scope, whether that is a city (e.g.\ somewhere in Paris), an island (typical of Malta), a country (difficult to be more specific than Thailand), a group of countries (Baltic states), or a broader region (e.g.\ any tropical island).

  \item The conclusion does not need to restate the evidence. Redundancy should be avoided, though briefly restating a key factor is acceptable if it naturally supports the final judgment. For example:
  \begin{quote}
    ``This could be South Africa or Australia, but based on Z, South Africa is slightly more likely.''
  \end{quote}
\end{itemize}
    \end{quotebox}
    \caption{Guidelines for writing the conclusion in the reasoning chain}
    \label{box:conclusion_guidelines}
\end{figure*}

We list the guidelines we shared with our three human expert GeoGuessr players in \cref{box:reasoning_guidelines} and \cref{box:conclusion_guidelines}.

\subsection{Expert Reasoning Chain Example}

\begin{figure*}[h!]
    \centering
    \includegraphics[width=\textwidth]{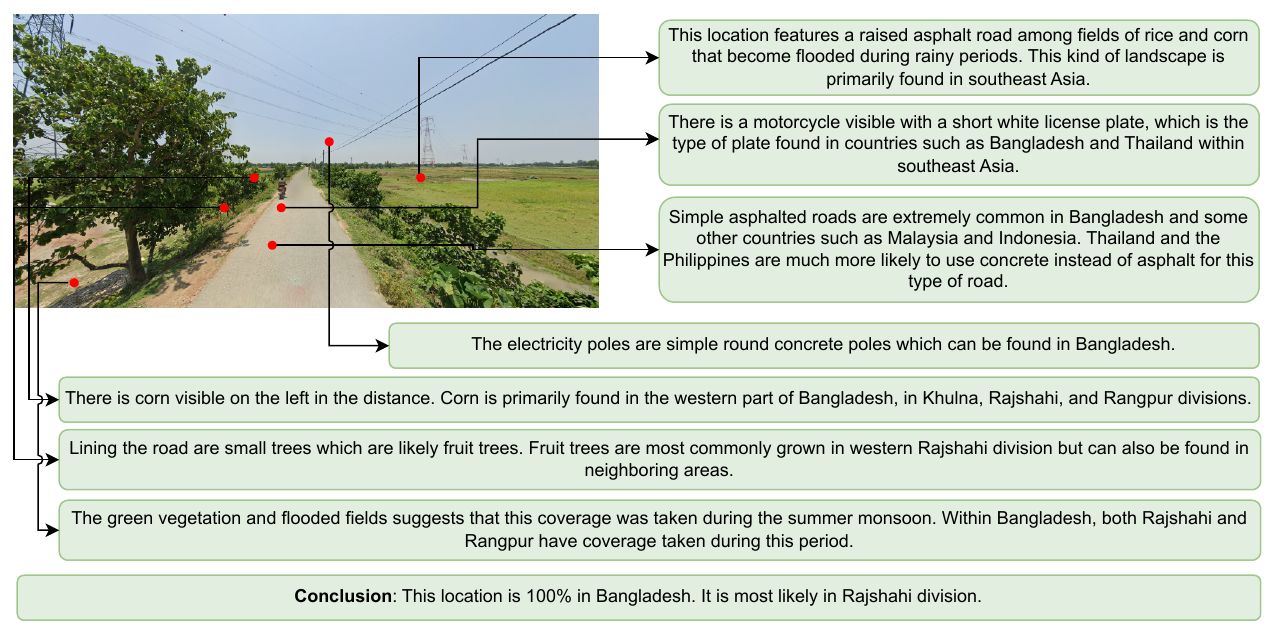}
    \caption{\textbf{Another Example of Expert Reasoning Chains.} Similar to \Cref{expert_chains}, this expert reasoning chain is non-exhaustive.}
    \label{expert_chains_appendix}
\end{figure*}

\Cref{expert_chains_appendix} shows another example of a geolocation reasoning chain written by a human expert. The steps progress from coarse to fine manner before concluding with a country guess. Expert chains are non-exhaustive as they can cite multiple scene attributes.
\subsection{Human Grading}
\subsubsection{Human Grading Example}
\begin{figure*}[h!]
    \centering
    \includegraphics[width=\linewidth]{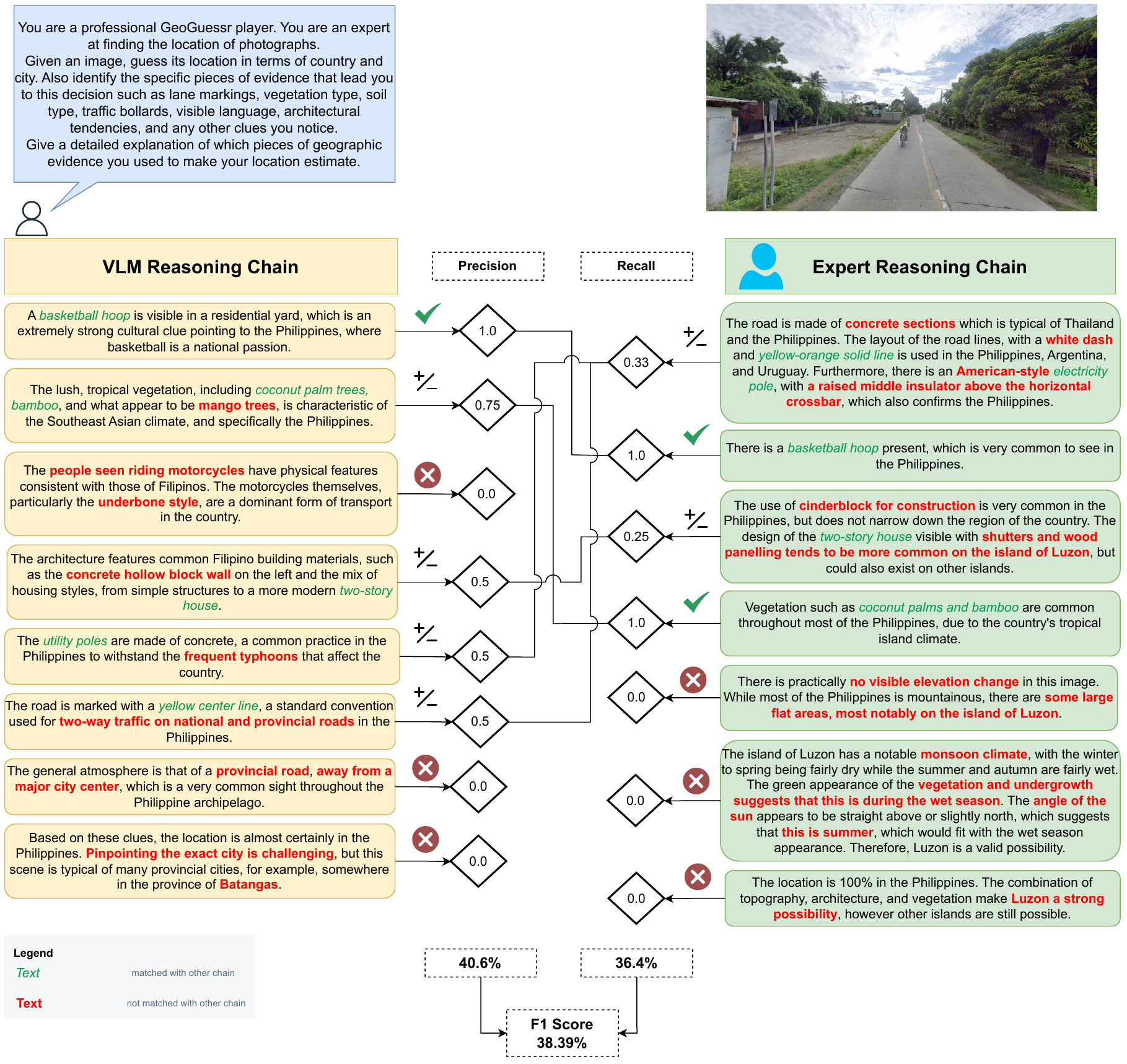}
    \caption{Example scoring of a candidate VLM chain using human grading. For each point in the VLM Reasoning chain, we compare it against the matching points on the Expert Reasoning Chain and assign a score. We repeat this process for each point on the expert chain. The overall precision and recall is the average of these scores. This is finally used to compute the F1 score.}
    \label{human_scoring_ex}
\end{figure*}

\Cref{human_scoring_ex} depicts how an expert human evaluator scores a candidate VLM reasoning chain with an Expert Reasoning Chain. The expert who scored it is different from the one who wrote the reasoning chain to avoid bias. 
\subsubsection{Guidelines}
\label{app:human_grading_guidelines}

\begin{figure*}[h]
    \centering
    \begin{quotebox}
    \begin{itemize}
        \item Each bullet point is scored some amount between [0,1] based on the amount of ``support'' from the other chain. All evidence in a single bullet point should be considered equally in the score.
        \item Bullet points can be supported by multiple bullet points in the other chain. It doesn’t need to be 1 to 1 matching. It can be 1 to many.
        \item Conclusion points are not considered
        \item Some reasoning chains are coarse-to-fine. Graders should be lenient when scoring. For example, one chain that says “Southeast Asian architecture” and a reference chain that says “This architecture could be anywhere in Thailand” are in agreement because that chain already narrowed it down to Thailand.
        \item Graders are judging the image properties, not the correctness of the geographic support of those properties. For example, ``White outer lines typical of Africa'' and “White outer lines typical of the Middle East” are in agreement. 
    \end{itemize}
    \end{quotebox}
    \caption{Human Grading Guidelines}
    \label{box:human_grading_guidelines}
\end{figure*}
This section details the guidelines adopted by the human experts to score the candidate reasoning chains with the reference reasoning chains written by the best expert. \ref{box:human_grading_guidelines} lists these guidelines.

\subsubsection{1-To-All LLM-as-a-judge}

\begin{figure*}[t]
    \centering
    \begin{quotebox}
        Provided below are two reasoning chains, the first following ``FIRST REASONING CHAIN:'' and the second following ``SECOND REASONING CHAIN:''. 
    These represent reasoning chains for determining the location of a photo, and these chains make statements about scene attributes such as infrastructure, architecture, 
    vegetation and agriculture, geology, topological features, cultural clues, vehicles, language signs, and use these attributes to reason about location. 
    The first reasoning chain has multiple statements/points separated by the "||" delimeter. 
    The second reasoning chain only contains one statement/point.
    Your task is to give a score out of 100 that reflects how similar the second reasoning chain is to any statement from the first reasoning chain. Note, similarity is defined as how much the statements logically supports, reinforces, and overlaps with in terms of geographical evidence. A score of 100 means that the second reasoning chain is a direct paraphrase of one of the points in the first reasoning chain.
    
    STRICT SCORING RULES:
    \begin{enumerate}
        \item Only award points if the SAME EVIDENCE TYPE appears in both chains.
        \begin{itemize}
            \item Evidence type means the category of observation (e.g., infrastructure, architecture, vegetation, agriculture, geology, topography, cultural clues, vehicles, signage, language, etc.).
            \item If the second chain discusses a different evidence type than all points in the first chain, the score MUST be exactly 0.
        \end{itemize}
        \item Do NOT give any credit for:
        \begin{itemize}
            \item Both mentioning the same country or region.
            \item Both being plausible, consistent, or geographically compatible.
            \item Both being true statements about the same place.
            \item The second chain making a correct or realistic claim for that country.
        \end{itemize}
        \item Score 100 ONLY if the second chain is a direct paraphrase or clear restatement of one of the points in the first reasoning chain, using the SAME evidence type and SAME logical reasoning.
        \item Score 0 if:
        \begin{itemize}
            \item The evidence type differs (for example, one chain is about roads/infrastructure and the other is about vegetation or topography).
            \item The overlap is purely thematic (for example, both mention the Philippines but discuss different kinds of evidence).
            \item There is no direct logical support or shared evidence category.
        \end{itemize}
        \item Intermediate scores (1–99) should ONLY be given when:
        \begin{itemize}
            \item Both chains clearly use the same evidence type (e.g., both describe vegetation or both describe architectural style),
            \item AND the second chain partially supports or overlaps with a point in the first chain but is not a perfect paraphrase.
        \end{itemize}
        \item Never assign a nonzero score purely because both reasoning chains refer to the same country, region, or overall conclusion.
    \end{enumerate}
    Please respond \textbf{only} with a JSON object. Make sure you do not include any commentary or explanation. Example format: {``precision'': <score>}. FIRST REASONING CHAIN: SECOND REASONING CHAIN:
    \end{quotebox}
    \caption{LLM-as-a-judge Prompt}
    \label{box:llm_judge}
\end{figure*}

\label{app:pseudocode}
\begin{algorithm}[h] 
\SetAlgoLined 
\KwData{Chain 1, Chain 2, LLM} 
\KwResult{Similarity Score} 

initialize prompt to LLM judge with rubrics\; 

Scores = empty list\;
\For{each statement in chain 1}
{
    Add statement to prompt\;
    Add chain 2 to prompt\;
    response = LLM(prompt)\;
    Add response to Scores\;
}
Similarity Score = average(Scores) \;
\Return{Similarity Score}
\caption{One-to-all }
\label{one_vs_all_llm}
\end{algorithm}

The prompt for scoring one candidate point in the reasoning chain to the complete ground truth reasoning chain by the LLM judge is shown in \cref{box:llm_judge}. The complete algorithm is presented in \cref{one_vs_all_llm}.

\subsubsection{Key Points Guided LLM-as-a-judge}

\begin{figure*}[h]
    \centering
    \begin{quotebox}
        Provided below is a statement describing one or more scene attributes such as infrastructure, architecture, vegetation and agriculture, geology, topological features, cultural clues, vehicles and language signs that is used to reason about location.
    
    Your task is to only output a comma separated list of words or phrases from the statement that most accurately summarize the statement.
    
    Following is the statement: 
    
    Begin your answer with LIST:
    \end{quotebox}
    \caption{Prompt for Key Points Guided LLM-as-a-judge approach for scoring reasoning chains}
    \label{box:prompt_kp_llm}
\end{figure*}

\begin{figure*}[h!]
    \centering
    \includegraphics[width=\textwidth]{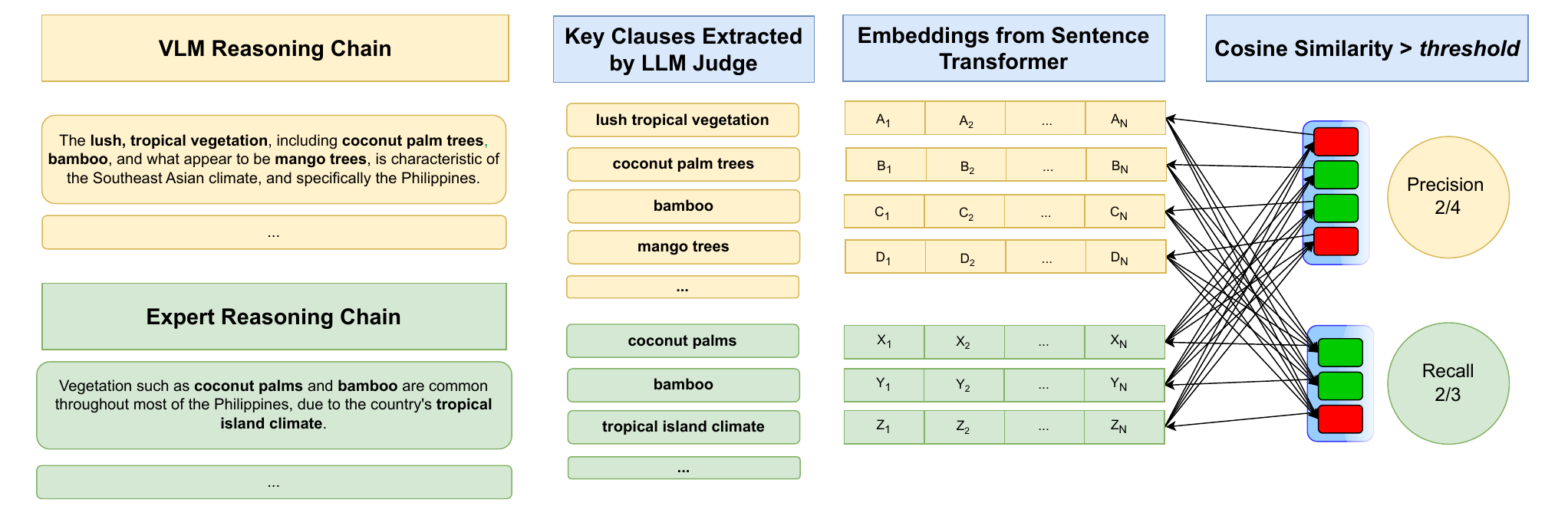}
    \caption{Key Points Guided LLM-as-a-judge Evaluation}
    \label{fig:kp_scoring}
\end{figure*}

\begin{algorithm}[h] 
\SetAlgoLined 
\KwData{Chain 1, Chain 2, LLM, Sentence Transformer} 
\KwResult{Similarity Score} 
Key Prompt = Initialize prompt to LLM judge for extracting atomic key points\;

Keys 1 = LLM(Key Prompt + Chain 1)\;
Keys 2 = LLM(Key Prompt + Chain 2)\;
Embeddings 1 = Sentence Transformer(Keys 1)\;
Embeddings 2 = Sentence Transformer(Keys 2)\;

\For{each key k1 in Keys 1}
{
    \For{each key k2 in Keys 2}
    {
        v1 = Get embeddings for k1 from Embeddings 1\;
        v2 = Get embeddings for k2 from Embeddings 2\;
        similarity = Compute Cosine Similarity between v1 and v2\;
        
        \If {similarity <= \textit{lower threshold} } {
            score = 0.0 \;
        }
        \If {similarity >= \textit{upper threshold} } {
            score = 1.0\;
        }
        \Else {
            score = (similarity - \textit{lower threshold})/(\textit{upper threshold} - \textit{lower threshold})
        }
        if current score is maximum for k1, store it\;
    }
}
Similarity Score = Weighted Sum of scores for keys\;
\Return{Similarity Score}
\caption{Key Points guided LLM judging}
\label{key_points_llm}
\end{algorithm}

The prompt used for converting the text to key points is shown in \cref{box:prompt_kp_llm}. \cref{fig:kp_scoring} shows the approach alongside the complete algorithm presented in \Cref{key_points_llm}. The hyperparameters for the thresholds that worked best were 0.45 and the sentence transformer used for this was \cite{sentence_transformer}.

\subsubsection{VLM-as-a-judge}

\begin{figure*}[h]
    \centering
    \begin{quotebox}
        Provided below is a reasoning bullet point for determining the location of a photo. This point makes a statement about scene attributes such as infrastructure, architecture, vegetation and agriculture, geology, topological features, cultural clues, vehicles, language signs and use these attributes to reason about location. 
    
    Your task is to determine if the bullet point is TRUE or FLASE based on the provided image. 
    
    TRUE means that all the scene attributes that are represented by the bullet point are present in the image. 
    
    Important Rules to Follow:
    \begin{itemize}
        \item Statements guessing the location such as "This location is in <country>" or "This location is in <city>" are ALWAYS TRUE.
        \item Do NOT judge it as TRUE or FALSE based on the guess of the location stated in the bullet point. For example, if the bullet point is "This vegetation is typically found in <country>", but the image is actually from a different country, it is TRUE as long as the vegetation is verifiable from the image.
        \item Statements about Generation coverage and the camera technology that cannot be verified from the image should always be considered as TRUE.
        \item If a bullet point clearly contradicts the image, it should be marked as FALSE. For example, if it states that there are "Yellow lane markings" but the image shows "White lane markings", this should be labelled as FALSE.
    \end{itemize}
    
    OUTPUT ONLY A SINGLE WORD TRUE or FALSE.
    
    Following is the reasoning bullet point: 
    \end{quotebox}
    \caption{Prompt for VLM-as-a-judge}
    \label{box:prompt_vlm_judge}
\end{figure*}
    
\begin{figure*}[h]
    \centering
    \begin{quotebox}
        Provided below are two reasoning chains, the first following ``GROUND TRUTH REASONING CHAIN:'' and the second following ``CANDIDATE REASONING CHAIN:''. 
    These represent reasoning bullet points for determining the location of a photo and these points make statements about scene attributes such as infrastructure, architecture, vegetation and agriculture, geology, topological features, cultural clues, vehicles, language signs and use these attributes to reason about location. 
    The ground truth reasoning chain is written by an expert who is always correct. 
    
    Your task is to compare the candidate bullet points to the ground truth and score it, out of 100, in terms of precision and recall. 
    Precision measures similarity of the candidate points to the ground truth reasoning chain. 
    Recall measures similarity of the ground truth bullet points to the candidate reasoning chain.

    Similarity is defined as how much the statements logically supports, reinforces, and overlaps with in terms of geographical evidence.
    A score of 100 means that the candidate reasoning chain is a direct paraphrasis of the points in the ground truth reasoning chain.
        
    Important Rules to Follow:
    \begin{itemize}
        \item Bullet points are the core unit that goes into the numerator and denominator of the recall and precision calculations. E.g. if there are 7 non-concluding, image-evidence bullet points, that will be the denominator. This is true even if 3 of the bullets are about road infrastructure.
        \item Each bullet point has equal amount of “support” from the other chain and contributes equally to the score out of 100.
        - Bullet points can be supported by multiple bullet points in the other chain. It does not need to be 1 to 1 matching. It can be 1 to many.
        \item Bullet points that are restating previous evidence should be disregarded (affecting neither the numerator or denominator). This goes beyond explicit “conclusion” bullets. Sometimes there are multiple conclusion-like bullet points. Sometimes there are bullet points that elaborate on the previous bullet points without pointing to new image evidence. Disregard these.
        \item ``White outer lines typical of Africa'' and ``White outer lines typical of Middle East'' are in agreement. We are judging the image properties, not the correctness of the geographic support of those properties.
        \item Only award points if the same type of evidence appears in both chains (example, both mention architecture, or both mention plant species).
        \item Do NOT award points for general overlap in country, theme, or plausibility.
        \item If the candidate chain uses a different evidence type (example, it discusses architecture, but the ground truth chain only talks about vegetation), the precision score must be 0 for this bullet point.
        \item Do not infer connections that are not explicitly stated.
        \item Do not award points simply because both chains arrive at the same conclusion (example, same country) if the evidence types are unrelated or not logically connected.
        \item The reasoning chains should be logically consistent. If candidate reasoning chain introduces evidence that contradicts or is irrelevant to the other, score 0 for precision.
    \end{itemize}
        
    OUTPUT ONLY A JSON BLOB representing these two values as follows: { ``precision'': , ``recall'': }
    
    GROUND TRUTH REASONING CHAIN:
    CANDIDATE REASONING CHAIN:
    \end{quotebox}
    \caption{Prompt for the LLM judge within the VLM-as-a-judge approach for evaluation}
    \label{box:llm_vlm_judge_prompt}
\end{figure*}

\begin{algorithm}[h] 
\SetAlgoLined 
\KwData{Candidate Chain, Reference Chain, Image, LLM, VLM} 
\KwResult{F1 Score} 

Correctness Prompt = initialize prompt to VLM judge with rubrics\; 
True Count = 0 \;
\For{each statement in Candidate Chain}
{
    Add statement to Correctness Prompt\;
    response = VLM(Correctness Prompt, Image)\;
    \If{\textit{true} in response} {
        True Count += 1\;
    }
}
Correctness = True Count / Number of Statements in Candidate Chain

Scoring Prompt = initialize prompt to LLM judge with rubrics\;
Add Reference Chain to Scoring Prompt\;
Add Candidate Chain to Scoring Prompt\;
Precision, Recall = LLM(Scoring Prompt) \;
Precision *= Correctness\;
Compute F1 score using Precision and Recall\;
\Return{F1 Score}
\caption{VLM-as-a-judge}
\label{vlm_judge}
\end{algorithm}

The prompt we used to get the correctness score from the VLM is shown in \cref{box:prompt_vlm_judge} and the prompt to the LLM is shown in \cref{box:llm_vlm_judge_prompt}. The complete algorithm is presented in \cref{vlm_judge}.

\subsection{Judging Methods}

\begin{table}[!ht]
\centering

\begin{adjustbox}{max width=\linewidth}
\begin{tabular}{lccc}
\toprule
\textbf{Experiment} & \textbf{One-To-All} & \textbf{KeyPoint} & \textbf{VLM-based} \\
\midrule
Spearman Correlation Coefficient & 0.6673 & \textbf{0.6708} & 0.4192 \\
Kendall Correlation Coefficient & 0.4890 &  \textbf{0.4953} & 0.2892 \\
\bottomrule
\end{tabular}
\end{adjustbox}
\label{tab:corr_coeffs}
\caption{Evaluation of Judging Methods using Kendall and Spearman Correlation Coefficients. The KeyPoint just narrowly performs better than One-To-All approach. However, the latter still remains our choice given its lower mean absolute error and higher Pearson correlation coefficient.}
\end{table}

\begin{figure*}[h]
    \centering
    \includegraphics[width=\textwidth]{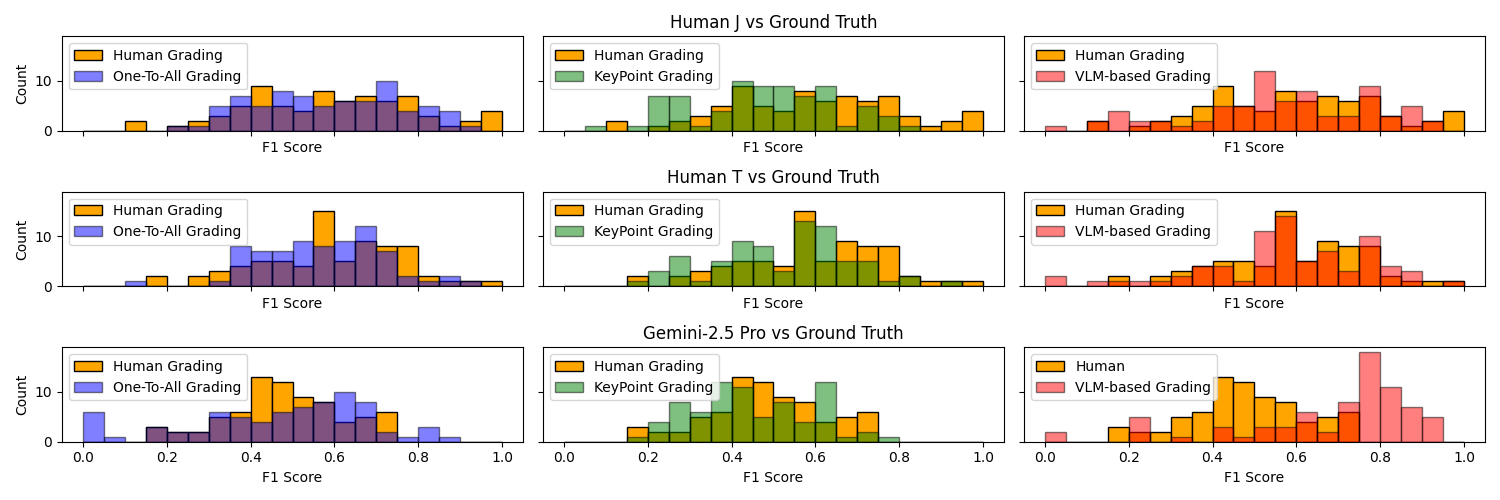}
    \caption{Mean Absolute Error of judging methods across three different experiments}
    \label{judging_metrics}
\end{figure*}

In \Cref{judging_metrics} we observe that the histograms for the count of the F1 scores best align with the one-to-all scoring approach across the three different flavors of the candidates. 

\begin{table}[!ht]
\centering
\label{tab:ICC}
\begin{adjustbox}{max width=\linewidth}
\begin{tabular}{lcc}
\toprule
\textbf{Metric} & \textbf{ICC3} & \textbf{ICC3k} \\
\midrule
Precision & 0.708294 & 0.879290 \\
Recall & 0.721981 & 0.886242 \\
F1 & 0.787125 & 0.917306 \\
\bottomrule
\end{tabular}
\end{adjustbox}
\caption{Inter Annotator Agreement ICC values for Precision, Recall and GeoReasoning F1 scores}
\end{table}

\begin{figure*}[h]
    \centering
    \includegraphics[width=\textwidth]{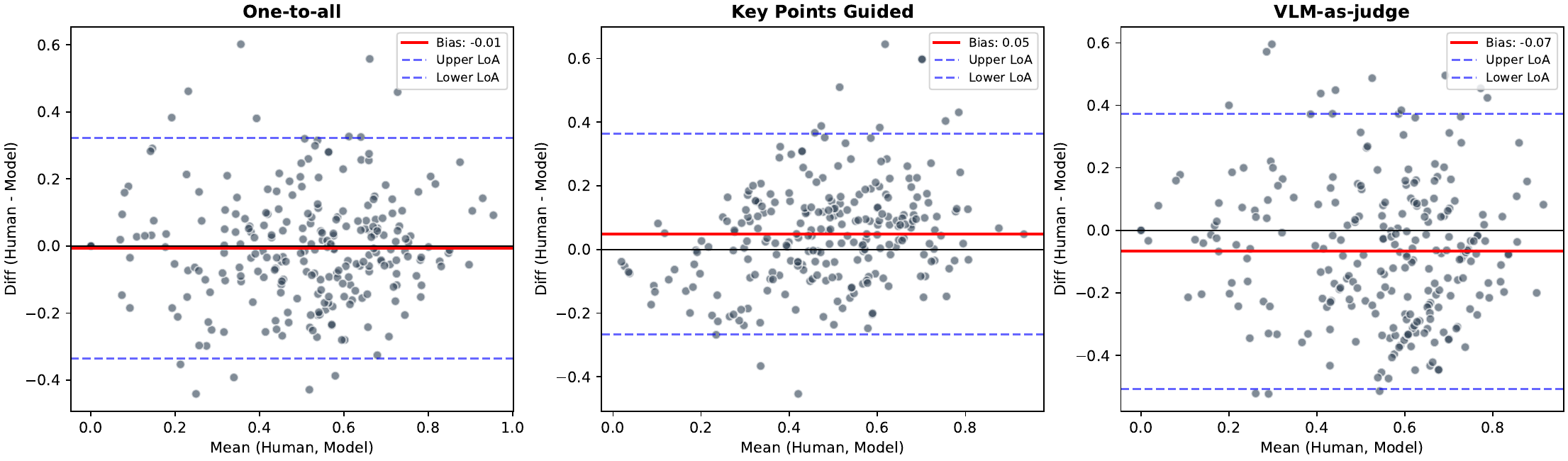}
    \caption{Bland Altman Plots for the Judging Methods}
    \label{bland_altman}
\end{figure*}

In \Cref{bland_altman} we found the mean difference line to be closest to zero for the one-to-all LLM judge, with a bias of 0.00 compared to 0.05 and -0.07 for key points-guided and VLM-as-a-judge, respectively. This reinforces our choice of the one-to-all LLM judge for our benchmark
\subsection{Results}

\begin{table*}
\centering

\begin{adjustbox}{max width=\textwidth}
\begin{tabular}{l c c c}
\toprule
\textbf{Candidate} & \textbf{N} & \textbf{95\% CI for F1} & \textbf{Pearson coeff.} \\ 
\midrule

\multicolumn{4}{c}{\textbf{Human Expert}} \\
\midrule
Human Expert Average & 450 & \textbf{[55.12, 58.26]} & -0.1135 \\

\midrule
\multicolumn{4}{c}{\textbf{Open-weight VLMs}} \\
\midrule
Llama-3.2-11B-Vision-Instruct \shortcite{llama3herdmodels} & 800 & [22.86, 25.14] & -0.1388 \\
Qwen2.5-VL-7B-Instruct \shortcite{qwen25vltechnicalreport} & 800 & [28.03, 30.15] & -0.2530 \\
Qwen3-VL-8B-Instruct \shortcite{qwen3vltechnicalreport}    & 800 & [21.17, 23.65] & 0.0336 \\
Gemma-3-12b-it \shortcite{gemma3technicalreport}           & 800 & [24.05, 26.42] & -0.0506 \\

\midrule
\multicolumn{4}{c}{\textbf{Proprietary VLMs}} \\
\midrule
GPT-5 \shortcite{openai2025gpt5}                           & 800 & [40.73, 43.88] & -0.1833 \\
GPT-4.1 \shortcite{openai2025gpt41}                        & 800 & [42.34, 45.14] & -0.0538 \\
Gemini-3-Pro \shortcite{google2025gemini3}                 & 800 & [38.79, 41.60] & -0.1154 \\
Gemini-2.5-Pro \shortcite{gemini25}                        & 800 & [41.99, 44.84] & 0.0109 \\
Gemini-2.5-Flash \shortcite{gemini25}                      & 800 & [41.34, 44.19] & 0.0651 \\
\bottomrule
\end{tabular}
\end{adjustbox}
\label{tab:model_statistics}
\caption{Scoring Statistics across Candidate Reasoning Chains. The confidence intervals for each candidate are tight and the Pearson Correlation coefficient values for Length of Chain and F1 score clearly shows no correlation between these factors.}
\end{table*}

\subsubsection{Experimental Setup}

We use the default temperature setting of 0.7 for QwenVL and Llama, 1.0 for Gemma, while for max new tokens, we set it to 512. For Qwen 3 VL specifically, we set the max new tokens to 2048 to avoid incomplete reasoning chains. When using VLM-as-a-judge, we set the max new tokens to a smaller value (10) to optimize runtime, while the LLM component has a max new tokens of 2048. While generating candidate chains from VLMs such as Qwen-2.5 and Qwen-3, this value for max new tokens was insufficient and the output did not complete. For these, we increased this value to 32768 to allow their chains to complete until the end of token.
\begin{figure*}[h]
    \centering
    \begin{quotebox}
        You are a professional GeoGuessr player. You are an expert at finding the location of photographs. 
Given an image, guess its location in terms of country and city. Also identify the specific pieces of evidence that lead you to this decision such as lane markings, vegetation type, soil type, traffic bollards, visible language, architectural tendencies, and any other clues you notice. 
Give a detailed explanation of which pieces of geographic evidence you used to make your location estimate. 
Output each point on a separate line for the reasoning chain and only generate as output the reasoning chain without annotations and titles.
    \end{quotebox}
    \caption{Prompt provided to the VLM candidates to induce them to generate reasoning chains}
    \label{box:vlm_gen}
\end{figure*}
The prompt used for generating the reasoning chains used commonly for all the VLM candidates is shown in \cref{box:vlm_gen}.

\begin{figure*}[h]
    \centering
    \includegraphics[width=\textwidth]{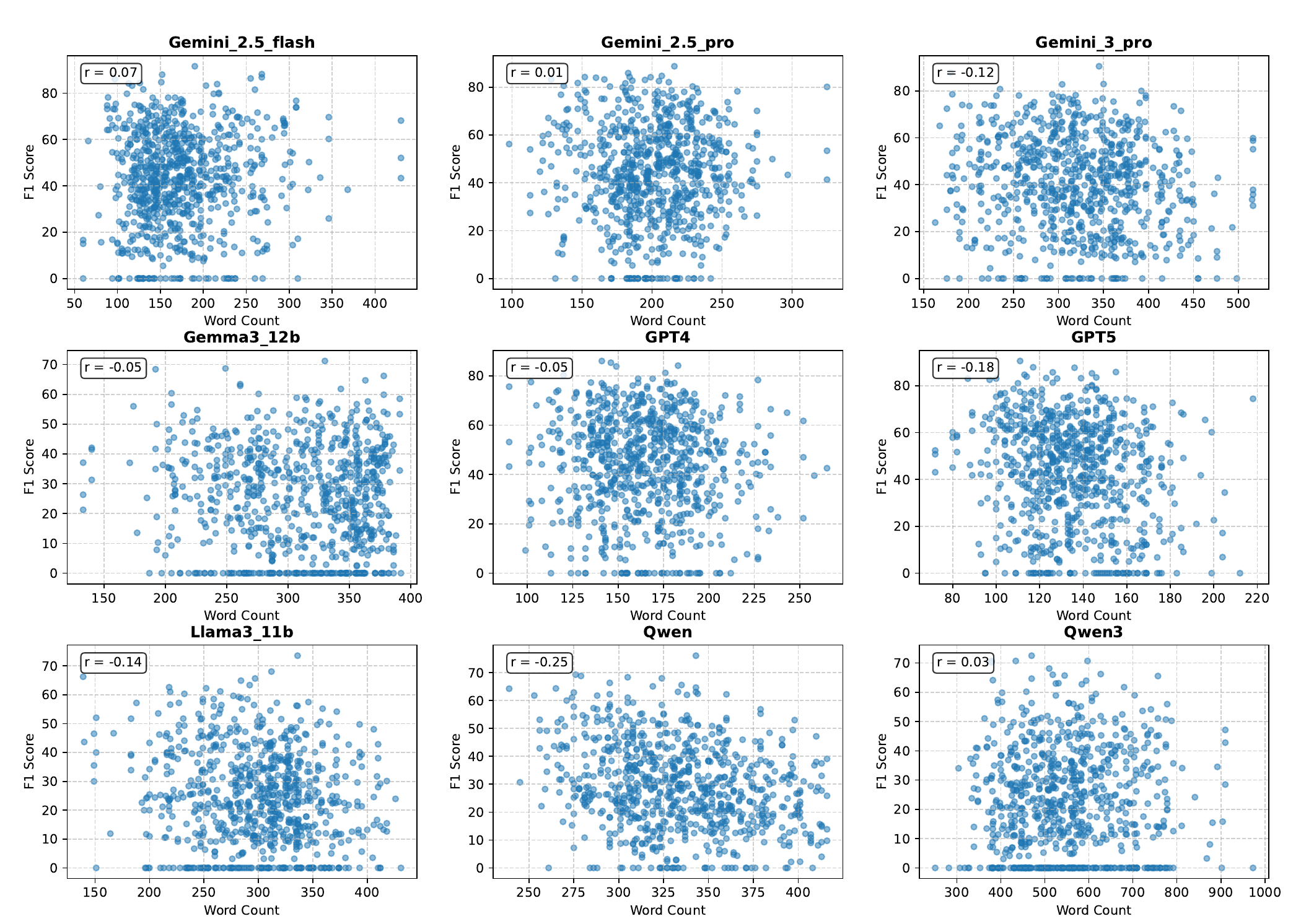}
    \caption{Distribution of F1 scores versus chain length measured by the number of words for multiple candidates clearly showing no correlation between the score from the one vs all approach and the number of words in the chain.}
    \label{f1_chain_length}
\end{figure*}

\begin{figure*}[h]
    \centering
    \includegraphics[width=\linewidth]{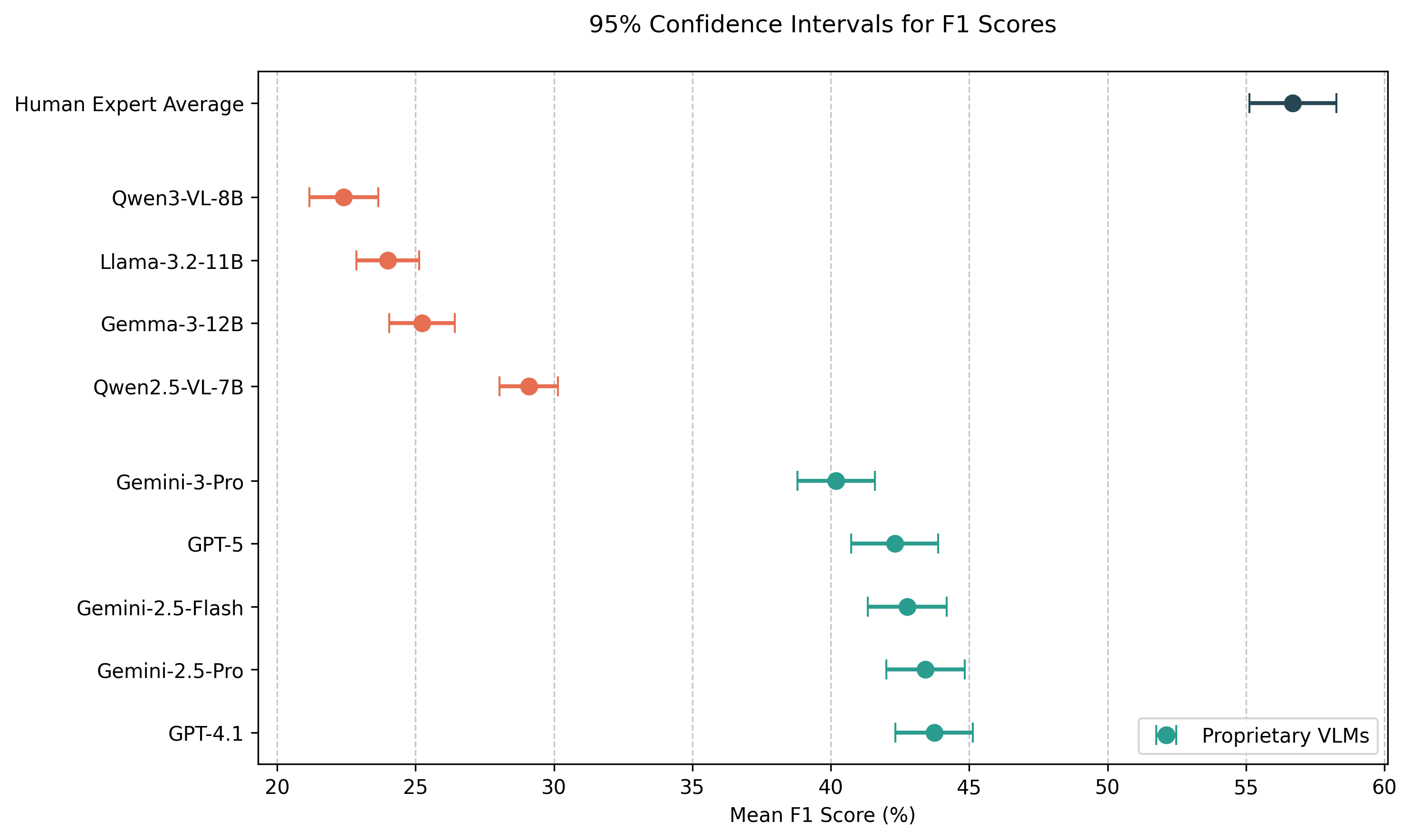}
    \caption{Graph showing confidence intervals for each candidate VLM}
    \label{confidence_intervals}
\end{figure*}

Additional statistics surrounding the candidates evaluated is shown in table \Cref{tab:model_statistics}. The confidence intervals for each candidate are narrow with the start and end of each interval being close to each other. Furthermore, we do not observe any correlation between the length of the chain (taken by the number of words) and the F1 score for any candidate thereby further strengthening our benchmark's evaluation approach. \Cref{f1_chain_length} shows the graphical distribution capturing the same result of these two measures being uncorrelated and \Cref{confidence_intervals} shows the confidence intervals in a graphical representation.

\subsubsection{Qualitative Results}
\begin{figure*}[h!]
    \centering
    \includegraphics[width=\textwidth]{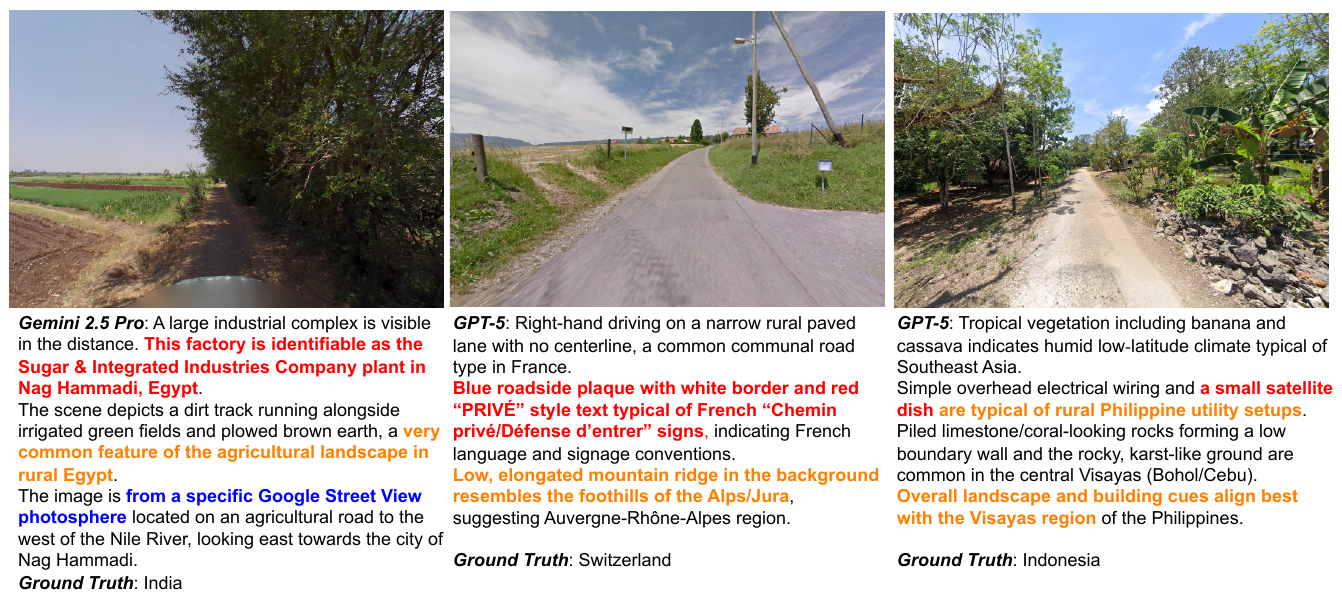}
    \caption{Text highlighted in \textbf{\textcolor{orange}{orange}} shows geographic misattribution where the scene attribute despite being in multiple countries is associated with a specific country. Text highlighted in \textbf{\textcolor{red}{red}} shows a hallucination where the referred scene attribute is absent and is not corroborated by the image. Text highlighted in \textbf{\textcolor{blue}{blue}} shows a red herring where an irrelevant topic is introduced completely out of context. Text highlighted in \textbf{\textcolor{MyPurple}{purple}} depicts an axiomatic irrelevance which is an obvious statement made about a geolocation attribute that is not contributing to the reasoning}
    \label{common_error_scenarios_appendix}
\end{figure*}

\Cref{common_error_scenarios_appendix} shows more examples of failure scenarios similar to those shown in \Cref{fig:common_error_scenarios}.

\begin{figure*}
    \centering
    \includegraphics[width=\textwidth]{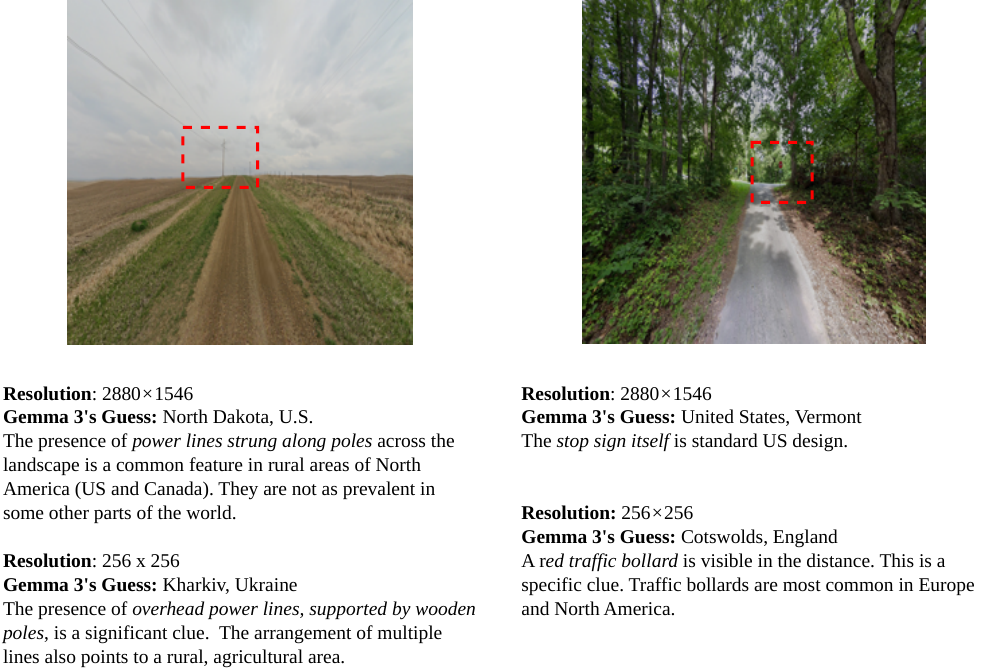}
    \caption{Down-sampling the input image passed to VLMs such as Gemma-3 causes evidence attributes to be excluded in its reasoning chain. Despite these still being visible to the human eye in the down-sampled image, these are not cited. Therefore, further strengthening our claim of lossy input image encodings.}
    \label{ablation_pixel_space}
\end{figure*}

We performed another ablation experiment to further strengthen our claim about the failure in the vision module of VLMs to refer to evidence attributes that occupy a smaller pixel space. \Cref{ablation_pixel_space} shows the results of the ablation experiment. In both examples, geographic evidences such as power lines and road signs are still clearly visible to human experts but the output reasoning chain contains mis-attributions to the geographic attributes thereby resulting in the overall guess to be completely changed to a different location.

\end{document}